 \documentclass[conference]{IEEEtran}

\usepackage[skip=2pt,font=small]{caption}
\usepackage{graphicx} % Required for inserting images
\usepackage{xspace}
\usepackage{algorithm}
\usepackage{tablefootnote}
\usepackage{algpseudocode}
\usepackage{adjustbox}
\usepackage{threeparttable,booktabs}
\usepackage{siunitx}
\usepackage[version=4]{mhchem} 
\usepackage{amsfonts}
\usepackage{pifont}
\usepackage{multirow}
\usepackage{graphicx}
\usepackage{subcaption}
\usepackage{xcolor}
\usepackage{tabularx} 
\usepackage{booktabs}
\usepackage{multicol}
\usepackage{enumitem}
\usepackage{lipsum} % for dummy text
\usepackage{cite}
\usepackage{url}

\usepackage{amsmath}
\usepackage[dvipsnames]{xcolor}
\usepackage{tikz}
\usetikzlibrary{tikzmark,arrows.meta,calc,positioning}

\newcommand{\mnode}[3][]{\tikzmarknode[#1]{#2}{#3}}

\tikzset{
  annot/.style = {-{Latex[length=2mm]}, line width=0.3pt},
  lab/.style   = {font=\scriptsize, align=center}
}

\usepackage[normalem]{ulem}

\newcommand{\fram}{\textsc{Lumos}\xspace}

\begin{document}

% \title{Loxia: A Self-Guided System for Optimizing Feature Selection and Model Pruning in Scientific Machine Learning Models\vspace{-6mm}}
\title{\fontsize{19pt}{19pt}\selectfont \fram: Democratizing SciML Workflows with L0-Regularized Learning for Unified Feature and Parameter Adaptation}
\vspace{-6mm}
% \title{\fontsize{20pt}{20pt}\selectfont \fram: L0-Regularized Learning that Folds Scientific Feature Selection and Parameter Pruning into SciML Models}

\author{\IEEEauthorblockN{Shouwei Gao$^{\dagger}$, Xu Zheng$^{*}$, Dongsheng Luo$^{*}$, Sheng Di$^{\ddagger}$, Wenqian Dong$^{\dagger}$}
\IEEEauthorblockA{
$^{\dagger}$Oregon State University, $^{*}$Florida International University, $^{\ddagger}$Argonne National Laboratory
}}

\maketitle

\begin{abstract}
% The rapid growth of scientific machine learning (SciML) models has accelerated discovery across scientific domains. However, the application of SciML models encounters three primary challenges: (1) identifying the most pertinent input features, (2) managing the complexity of overly large models that impede their practical application, and (3) addressing high computational demands. We introduce \fram, an end-to-end, self-guided framework designed to address these challenges through efficient feature selection and model compression. \fram leverages semi-stochastic gating and reparameterization techniques to dynamically prune both input features and model parameters during training. By eliminating redundant model parameters, \fram enhances computational and communication efficiency, improving throughput, latency, storage, and energy usage. 
% We evaluate \fram across diverse SciML workloads such as cosmology and molecular sciences, demonstrating its effectiveness and generalizability.
% Our evaluation of 12 SciML models shows that \fram achieves a 71.71\% reduction in parameters, a 7.58$\times$ speedup, and a 50.69\% reduction in energy consumption on average. We employed Distributed Data Parallel (DDP) training on up to eight GPUs to demonstrate the scalability of the \fram while ensuring that the training overhead remains negligible.
\textcolor{black}{
The rapid growth of scientific machine learning (SciML) has accelerated discovery across diverse domains, yet designing effective SciML models remains a challenging task. In practice, building such models often requires substantial prior knowledge and manual expertise, particularly in determining which input features to use and how large the model should be. We introduce \fram, an end-to-end framework based on L0-regularized learning that unifies feature selection and model pruning to democratize SciML model design. By employing semi-stochastic gating and reparameterization techniques, \fram dynamically selects informative features and prunes redundant parameters during training, reducing the reliance on manual tuning while maintaining predictive accuracy. We evaluate \fram across 13 diverse SciML workloads, including cosmology and molecular sciences, and demonstrate its effectiveness and generalizability. Experiments on 13 SciML models show that \fram achieves 71.45\% parameter reduction and a 6.4× inference speedup on average. Furthermore, Distributed Data Parallel (DDP) training on up to eight GPUs confirms the scalability of \fram.}
\end{abstract}
% \renewcommand\IEEEkeywordsname{Keywords}
% \begin{IEEEkeywords}
% Scientific Machine Learning, Feature Engineering, Model Compression

% \end{IEEEkeywords}

% %%
% %% This command processes the author and affiliation and title
% %% information and builds the first part of the formatted document.
% \settopmatter{printfolios=true}

% \maketitle

\thispagestyle{plain}
\pagestyle{plain}

\section{Introduction}

%\textbf{Background.} 
% Scientific applications need to encompass a vast landscape of computational tasks to resolve non-trivial scientific problems.
Scientific applications span a wide spectrum of computational tasks aimed at solving complex, real-world problems. However, these tasks are often very computation-intensive due to the high complexity of simulation algorithms and data analysis. %For example, leveraging high-performance computing (HPC) for large-scale power grid simulations is essential to meet latency-sensitive demands and ensure national power system stability; however, the underlying data-dependent simulation is currently difficult to parallelize efficiently.
%
%The Eastern Interconnection power system has over 70,000 electrical nodes (also known as ``buses''), and a one-time simulation on this power system must consider thousands of daily security planning scenarios (i.e., all possible grid failure cases). Each scenario represents a (non-)linear programming problem involving hundreds of thousands of differential-algebraic equations, with computation times ranging from several microseconds to minutes~\cite{wu2021economic, messina2017exascale}. 
%Current research shows that employing emerging parallel devices, such as multi-core CPUs, or multi-threading GPUs, fails to offer performance gains, due to some data-dependent but computation-dominant parts such as LU matrix factorization in traditional solvers ~\cite{davisUserGuideKLU2022}.
%
Scientific Machine Learning (SciML) models can address these challenges with their highly accurate modeling capabilities on complex (non-)linear scientific datasets, concurrent computation patterns on AI/ML blocks, and less sensitivity to processor architectures. Recent studies demonstrate that integrating SciML models with traditional HPC applications can achieve significant speedups of $10^3$ or more~\cite{panayides2020ai,brace2022coupling,clyde2023ai}. SciML has also led to breakthroughs in diverse fields, gaining widespread adoption such as improving predictions of extreme weather events~\cite{rasp2018deep} in climate science; enhancing gene expression analysis~\cite{eraslan2019single} in genomics; and accelerating the identification of potential drug candidates~\cite{stokes2020deep} in drug discovery, respectively. 
Meanwhile, SciML is undergoing a paradigm shift with the advent of foundation models like AlphaFold-3~\cite{desai2024review}, PanGu-$\Sigma$~\cite{ren2023pangu}, and ClimaX~\cite{nguyen2023climax}, which are pretrained models on large and general datasets and can be adapted to various downstream tasks through fine-tuning \cite{qiao2024state,nakata2023end,baek2024accurate,townshend2021geometric,jiang2021interactiongraphnet,jiang2022predicting,corso2022diffdock,stark2022equibind,liao2019deepdock,lu2022tankbind,zhou2023uni,shen2022e2efold,van2010pushing}. These foundation models has strong generality due to trained from boarder dataset, and cost-efficient with less training steps.
% However, in both mainstream SciML workflows, the efficiency of scientific input features and model parameters remains poorly studied. As shown in Figure~\ref{xxx}, a domain scientist would like to build a SciML model, with a collected scientific dataset (i.e., SciML inputs and outputs) on hand. There are two practical scenarios: 

\begin{figure}[ht]
    \centering
    \includegraphics[width=\linewidth]{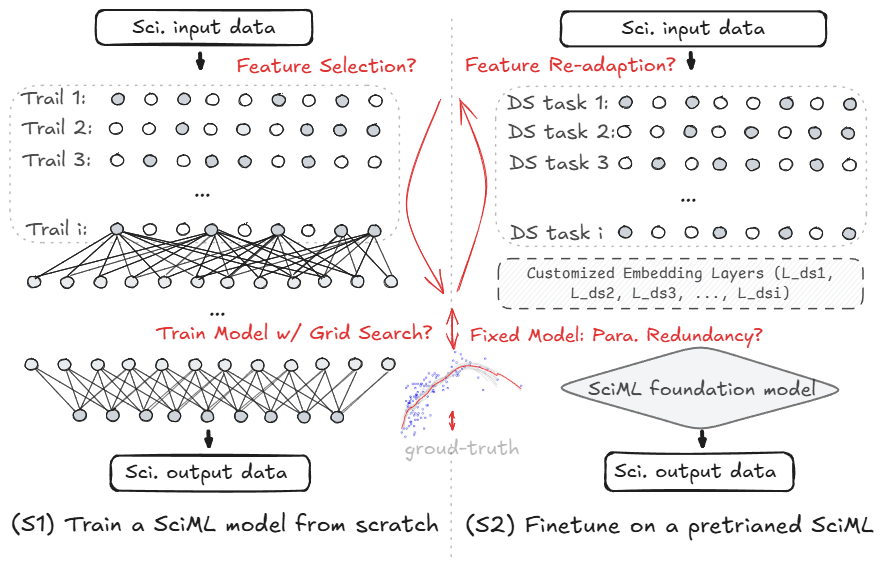}
    \caption{(S1) Training from scratch: domain scientists must select physics-informative scientific input features and co-design the model architecture and parameters, often relying on costly grid search and ad hoc heuristics. (S2) Fine-tuning a foundation model: the pretrained backbone is fixed, and adaptation is achieved by re-mapping input features and adjusting subsets of parameters. Both workflows face common bottlenecks in \textit{feature efficiency} and \textit{model/parameter efficiency}, motivating the development of unified frameworks such as \fram.}
    \label{fig:intro}
    % \vspace{-2em}
\end{figure}

However, in both mainstream SciML workflows, the efficiency of scientific input features and SciML model parameters remains poorly studied. Figure~\ref{fig:intro} showcases the two scenarios \textbf{(\underline{S1}} and \textbf{\underline{S2})} present critical and shared challenges in the designing pipelines. 

In the case of (\textbf{\underline{S1}}) training from scratch, domain scientists must decide the importance of scientific input features in the training of the SciML model. Identifying the most relevant input features is often difficult due to the complex, high-dimensional nature of scientific systems, where intricate variable interactions and hidden dependencies are common. Selecting salient features from a vast array of potential inputs is crucial for achieving accurate and efficient approximations~\cite{tripathy2018deep}, yet existing feature selection methods often suffer from low explainability, neglect of output quality, or excessive computational overhead (see Section~\ref{sec:background}.1). In practice, this problem is compounded by the need to co-design feature choices with the model architecture and hyperparameter search. Traditional training practices frequently resort to exhaustive grid search across architectures, parameters, and feature subsets~\cite{dong2023autohpcnet, wang2020fcos}. However, such brute-force approaches are labor-intensive and may result in significant redundancy, as not all input dimensions contribute equally to the predictive accuracy of the model. Also, this inefficiency is even worse when the model size grows and the parameter space becomes prohibitively large, making scalability a critical issue.

In contrast, \textbf{(\underline{S2})} fine-tuning a pre-trained foundation model shifts the burden from training an entire model to re-adapting customized embedding layers for the specific dataset at hand (based on the needs of downstreaming tasks). Here, the central question is not only which features are most relevant, but also how to best re-map scientific input descriptors into the latent representation space of the foundation model. While foundation models provide strong generalization ability, naïve adaptation can still result in parameter inefficiencies and performance bottlenecks, particularly when downstream datasets are small, noisy, or domain-shifted. With fixed and large pretrained weights, these models risk over-parameterization, leading to overfitting, high energy costs, and inference delays. Moreover, over-complex models consume large amount of memory and computational resources, conflicting with real-time simulation or analysis needs in scientific domains such as high-energy physics, where computational bottlenecks may hinder the detection of rare events~\cite{baldi2014searching}. Existing efficiency strategies, such as model compression, provide partial solutions but still face major limitations (see Section~\ref{sec:background}.2).

At their core, both scenarios highlight two intertwined and insufficiently addressed challenges: (i) \textit{feature efficiency}, that determines which scientific input features are most relevant and how to adapt them effectively across tasks and domains; ii) \textit{model architecture and parameter efficiency}, that decides how to allocate, adapt, or prune parameters to balance accuracy with computational cost, particularly under the constraints of heterogeneous scientific datasets.

To tackle these challenges, we propose \fram, an end-to-end, self-guided framework that applies L0-regularized learning that folds scientific feature selection and parameter optimization (such as structure pruning) into a traditional SciML training or fine-tuning pipeline. \fram enables domain scientists to construct compact but efficient models without labor-intensive manual intervention or domain-specific heuristics. By automating these optimization steps, \fram facilitates both from-scratch surrogate training and fine-tuning of foundation models, increasing stakeholders’ confidence in deploying SciML in real-world scientific workflows.

%\textbf{Key Contributions.} 
This paper makes the following contribution:

\begin{itemize} 

    \item We propose \fram, a unified framework that can conduct feature selection and parameter pruning simultaneously for SciML models. \fram leverages learnable and differentiable semi-stochastic neuron gates to enable dynamic optimization during training or fine-tuning.

    \item \fram introduces structural consistency mapper functions to track and reconciles architectural transformation to maintain the alignments of weights, activation tensors (such as KV tensors in attention layer) and outputs across SciML layers when conducting feature and parameter pruning. %This results in actual reductions in floating-point operations, translating into quantitatively computational gains.

    % \item \GS{\sout{\fram exposes learnable gating vectors applicable to all model components (e.g., input features, weights), providing interpretable indicators of component importance. It integrates lightweight tuning strategies that minimize the need for deep machine learning or domain-specific expertise, lowering the barrier to adoption in scientific applications.}
  \item \fram can be seamlessly integrated into existing SciML training or fine-tuning pipelines within their training phase, performing on-the-fly feature and parameter selection with negligible overhead.  
Across the 13 evaluated SciML models, the additional gate parameters introduced by \fram only take about 3\% on average.  
We demonstrate its effectiveness and generality across five representative layer types inlucidng fully connected layers, convolutional layers, Graph Isomorphism Network (GIN) layers, Graph Convolutional Network (GCN) layers, and attention layers, covering diverse scientific domains such as cosmology, molecular science, and materials research.
\end{itemize}

% \textbf{Experimental Evaluation.} We evaluated the framework across 12 diverse scientific applications and two simple showcase applications. Our experiments demonstrate that \fram achieves significant performance improvements across four key metrics: reduction of FLOPs, peak memory usage, speedup, and energy consumption. Additionally, we present case studies to showcase \fram's effectiveness in feature selection. We scale the training process across up to eight GPUs and demonstrate that the overhead introduced by \fram is negligible. \textcolor{red}{to Shouwei: double check and give some exact values for performance }

% \textcolor{green}{I didn't get the "Experimental Evaluation" here. Why should the experimental evaluation be here?}

\section{Background and Related Work}
\label{sec:background}

\subsection{Feature Selection}

Feature selection methods can be divided into three categories~\cite{chandrashekar2014survey}:   
\ding{182} \textit{Filtering}~\cite{peng2005feature} evaluates the correlation between features through statistical measurement.  Common techniques include Pearson's correlation coefficient \cite{biesiada2008feature}, Chi-square test \cite{jin2006machine}, and Mutual Information (MI) \cite{thang2010improved}. These methods rank features based solely on their intrinsic characteristics in relation to the target variable, making them computationally efficient. However, because they do not consider interactions between features and the model, filtering methods usually overlook complex dependencies within the data.
\ding{183} \textit{Wrapper methods} such as forward selection\cite{kamalov2024forward}, backward elimination\cite{park2025k}, or greedy search strategies\cite{pudjihartono2022review}, typically utilize iterative processes to select a subset of features. An example is the LM-FM framework proposed by Korfi et al. \cite{korfiatis2013classification}, which employs Local Maximization (LM) and Floating Maximization (FM) stages to optimize feature selection. While wrapper methods often deliver better performance than filter methods by considering feature interactions, they generally require computationally expensive manual tuning\cite{guyon2003introduction}.
\ding{184} \textit{Embedded methods} integrate feature selection directly into the model training process, often leveraging regularization techniques to identify and select relevant features. A well-known example is the Least Absolute Shrinkage and Selection Operator (LASSO)~\cite{tibshirani1996regression}, which enables input sparsification by adding a penalty as a coefficient on the model's loss function. Although effective, LASSO struggles with multicollinearity and non-differentiable optimization.

\subsection{Model Compression}
\label{sec:modelcompression}
%Model compression has been worked around to improve the model's efficiency without compromising accuracy\textcolor{red}{~\cite{xxx, xxx, xxx, xxx}}. Table~\ref{bechmarks} lists name a few. 
Model compression techniques aim to reduce the computational and memory footprint of deep learning models without significantly compromising prediction accuracy~\cite{cheng2017survey}. This is crucial for deploying complex models on resource-constrained devices and for reducing latency in time-critical scientific applications. %Table~\ref{tab:model_compression} presents a comparison of various model compression techniques, including pruning, lightweight design, knowledge distillation, and quantization. 

On the algorithm side, the community has addressed the computational complexity in machine learning models by pruning redundant and unnecessary model parameters to improve efficiency at high accuracy. \ding{182} \textit{Unstructured pruning }is individual weight pruning. 
For example, in a fully connected (FC) layer, unstructured pruning selectively removes certain weights, i.e., specific neuron connections with other layers. The importance of these connections determines their removal. For example, Lee et al. \cite{lee2019signal} proposed detecting insensitive weights using initialization conditions. %Recently, the magnitude-based pruning (MP) method uses the second derivatives of the loss function to evaluate weight importance. 
The unstructured pruning, however, has a critical drawback: it may result in sparse weight matrices with removed connections represented by zeros, which cannot inherently lead to compression or speedup without specialized hardware or libraries~\cite{zhou2021learning}.

% \begin{table}[!ht]
% \centering
% \caption{Comparison of SoTA model compression techniques. `-' means not supported.}
% \label{bechmarks}
% \footnotesize % Reduced font size
% \setlength{\tabcolsep}{3pt} % Reduce space between columns
% \begin{tabular}{l cccc}
% \toprule
%  % & \multicolumn{3}{c}{\textbf{Pros}} & \multicolumn{3}{c}{\textbf{Cons}} \\
% % \cmidrule(lr){2-4} \cmidrule(lr){5-7}
%  & \begin{tabular}[c]{@{}c@{}} Structured \\ computation\end{tabular} & \begin{tabular}[c]{@{}c@{}} Low tuning \\complexity\end{tabular} & \begin{tabular}[c]{@{}c@{}} Low accuracy \\degradation\end{tabular} & \begin{tabular}[c]{@{}c@{}}Hardware \\ compatibility\end{tabular}\\
% \midrule
% Unstructured Prun.   &  –         &   –          &  \checkmark &     –    \\
% Structured Prun.     & \checkmark &     –        &  –         & \checkmark\\
% Lightweight Design     & \checkmark &     –        &  –          & \checkmark\\
% Knowledge Distill. & \checkmark &   –         & \checkmark & \checkmark\\
% Quantization          & \checkmark & \checkmark  &    –      &   –       \\
% \textbf{\fram (Ours)}           & \checkmark & \checkmark  & \checkmark & \checkmark\\
% \bottomrule 
% \end{tabular}
% \label{tab:model_compression}
% \end{table}
Only a few leading semiconductor chips (such as NVIDIA A100/H100) have incorporated support for sparse operations.  
On hardware that lacks support for underlying sparse operators, which is commonly used by most domain scientists, the computation, such as sparse-dense matrix multiplication (SpMM), still operates in a dense (non-sparsified) manner.
\ding{183} \textit{Structured pruning}~\cite{ding2019centripetal, liu2021group, you2019gate} operates at the level of channel, neuron, or even layer. For example,  Li et al.~\cite{li2016pruning} employ the $L1$ norm, while He et al.~\cite{he2018soft} use the $L2$ norm on the filter and weights to guide the model pruning. Structured pruning provides a systematic reduction (by removing rows or columns) of computation matrices, offering flexibility in machine learning layers and portability across hardware architectures. However, most structured methods suffer from inefficiencies in both algorithmic performance and tuning complexity. For instance, one of the representative approaches~\cite{ye2018rethinking} relies on the hypothesis that ``smaller norm-less-informative" requires an extra sensitivity analysis for verification. This sensitivity analysis is an iterative optimization process, which is time-consuming and prone to getting trapped in local optima.  
Instead of pruning a proportional part of the model, \ding{184} \textit{lightweight design}~\cite{iandola2016squeezenet, howard2017mobilenets} and \ding{185}\textit{ knowledge distillation}~\cite{yu2019learning, tung2019similarity} compress the model by learning a new, smaller one. The lightweight design leverages a one-step training process, simplifying the tuning complexity but often sacrificing algorithmic performance~\cite{li2023model}. Knowledge distillation uses an advanced teacher-student framework that enhances algorithm performance but requires careful design and domain expertise for both models.

On the system side, diverse high-performance architectures, such as NVIDIA GPUs, Intel CPUs, AMD GPUs, Cerebras, SambaNova, and Graphcore, have advanced alongside the rapid growth of large-scale machine learning models. Vendors like NVIDIA, beginning with the DGX-2 (Volta) in 2017, introduced Tensor Cores to enable \textit{mixed-precision} and \textit{quantization} computation, significantly reducing computational complexity and memory usage. The latest H200 architecture supports a broad range of precision formats, including FP64, FP32, FP16, BF16, and INT8, and provides support for sparse matrix operations, further enhancing performance and efficiency.
While hardware-specific optimizations for quantization are beyond the scope of this paper and are left for future work, our proposed framework, \fram, offers a hardware-agnostic design for input and model sparsification. It can integrate with libraries across various architectures to explore and improve quantization efficiency. 

In contrast to the above methods, our proposed framework, \fram, automates and integrates input feature selection with model compression. \fram is hardware-agnostic, ensuring portability across various hardware architectures; it is self-guided, requiring no human supervision during the model tuning process; and it is algorithm-efficient, offering fine-grained control over the trade-off between compression ratio and performance.

\section{Design Overview}

\begin{figure}[ht]
    \centering
    \includegraphics[width=0.98\linewidth]{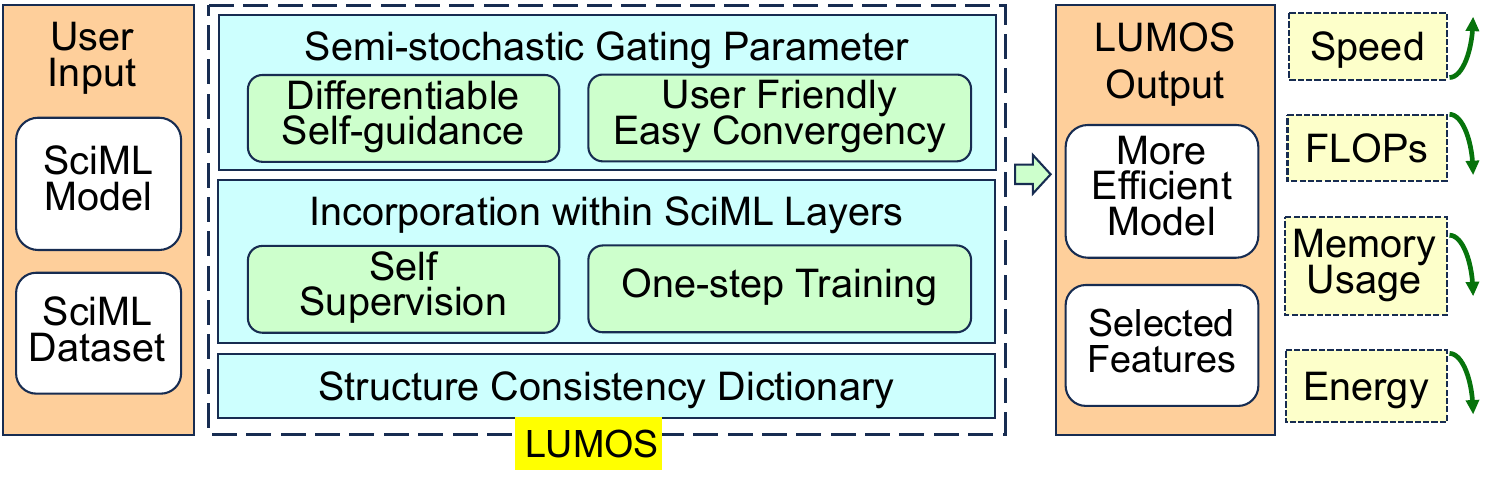}
    \caption{Overall workflow of \fram.} %\fram is an end-to-end framework that democratizes feature and parameter optimization process in SciML pipelines. \fram incorporates gating parameters into SciML layers to regularize structured pruning and ensure structural integrity, resulting in optimized models with improved computational and communication efficiency.}
    \label{overall}
    % \vspace{-2em}
\end{figure}

Figure~\ref{overall} gives an overview of \fram , an end-to-end framework to democratize feature selection, feature re-adaptation, and model archtecture search (i.e., structure pruning with parameter optimization) for SciML models. %\fram is self-guided 
\fram aims to tackle the broad and fundamental implications of SciML models, such as convolutional neural network (CNN), multilayer perception (MLP), graph neural network (GNN) and Transformer~\cite{vaswani2017attention}, while meeting the requirements of time and cost efficiency. 
Given a basic model and its dataset from users, \fram first applies gating parameters to each of the input features and model parameters (Section~\ref{sec:design}). Specifically, \fram adopts a semi-stochastic mechanism to regulate the initial states of the gates and applies a reparameterization function to make the gates differentiable within the backpropagation process.

Then, \fram incorporates the gates into SciML layers (Section~\ref{incorporation}). The cooperation regularizes the structured pruning on the vanilla structures of scientific inputs and SciML layers. Once integrated, the entire selection and optimization process in \fram is a one-step training (in \underline{\textbf{S1}}) or one-step fine-tuning (in \underline{\textbf{S2}}), without user supervision or domain knowledge.

Next, with the sparsity indication of gates, \fram incorporates a suite of structural alignment tools the \textit{Layer Consistency Mapper} and the \textit{Special Structure Coordinator} (Section~\ref{sec:scd}) when conducting model distillation, These tools maintain architectural integrity and ensure correct tensor alignment throughout the optimized model.
After the above offline phase, \fram produces an optimized model with the corresponding feature set, offering substantial improvements in computational and communication efficiency across latency, throughput, power, and storage in inference serving.%\footnote{The GitHub repository link of \fram will be included when applicable.} 
Code is available on GitHub.\footnote{\url{https://github.com/Picomp-lab/Loxia}}

\section{Problem Formulation}
\label{sec:problem}

\begin{figure}[htb]
    \centering
    \includegraphics[width=0.99\linewidth]{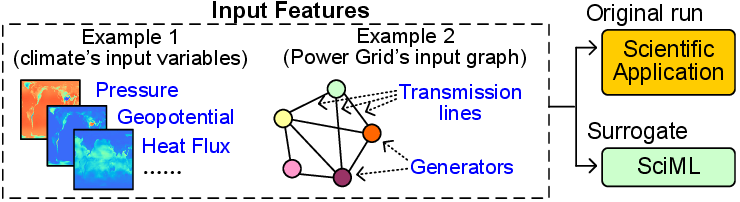}
    \caption{Illustration of Input Features: the blue words represent input features regarding two examples, climate application and power grid simulation, respectively.}
    \label{fig:features}
\end{figure}

A SciML model is fundamentally defined by two key components: model weights and input features (or feature datasets). The model weights are the learned parameters that constitute the trained model, typically organized across multiple layers. The input features (two examples are illustrated in Fig. \ref{fig:features}) represent the input parameters or variables specific to the scientific application.

For a SciML model associated with weights and feature datasets, the optimization can be defined as follows (The symbols and notations are listed in Table~\ref{Symbols} for reference throughout this paper):

\begin{table}[h]
\centering
\caption{Summary of Symbols}
\footnotesize
\label{Symbols}
\begin{tabular}{cl}
\toprule
\textbf{Symbol} & \textbf{Description} \\ \midrule
$\mathcal{F}$ & SciML model \\ 
$\mathcal{D}$ & Dataset  \\ 
$\mathbf{X}^{0}$ & Input features \\ 
$\mathbf{X}^{(l)}$ & Input to the $l$-th layer \\ 
$\mathbf{Y}^{gt}$ & Ground truth (output label) \\ 
$\mathcal{M}$ & Set of evaluation metrics \\ 
$l$ & Layer index \\ 
$\mathbf{H}^{(l+1)}$ & Output of the $l$-th layer \\ 
$\mathbf{W}^{(l)}$ & Weight matrix of the $l$-th layer \\
$m^{(l)}_{i}$ & The $i$-th gating parameter of $l$-th layer\\
$\mathbf{b}^{(l)}$ & Bias vector of the $l$-th layer \\ 
$\sigma$ & Activation function \\ 
$g(\mathbf{X}^{(l)}, \mathbf{W}^{(l)})$ & Layer-specific transformation function \\ \bottomrule
\end{tabular}
\end{table}

\noindent\textbf{Problem Formulation}.
Given a SciML model $\mathcal{F}$, a dataset $\mathcal{D} = {(\mathbf{X}^{0}, \mathbf{Y}^{gt})}$, and a set of evaluation metrics $\mathcal{M}$, the goal is to find an optimal subset of input features $\mathbf{X}'$$\subseteq$$\mathbf{X}^{0}$ and an optimized model $\mathcal{F}'$$\subseteq$$ \mathcal{F}$ that reduces the total number of parameters (weights) while preserving accuracy. The optimization can be expressed as:
\[
\arg\min_{\mathcal{F}' \subseteq \mathcal{F}, \mathbf{X}' \subseteq \mathbf{X}^{0}} \left| \mathcal{M}\left(\mathcal{F}(\mathbf{X}^{0}), \mathbf{Y}^{\text{gt}}\right) - \mathcal{M}\left(\mathcal{F}'(\mathbf{X}'), \mathbf{Y}^{\text{gt}}\right) \right|,
\]
where:
\begin{itemize}
    \item $\mathcal{F}(\mathbf{X}^{0})$ denotes the output of the base model using the full set of input features $\mathbf{X}^{0}$;
    \item $\mathcal{F}'(\mathbf{X}')$ denotes the output of the optimized model along with the selected subset of features $\mathbf{X}'$.
\end{itemize}

\section{Unifying Feature Selection and Model Compression}
\label{sec:design}

\subsection{Semi-Stochastic Gating Parameters} 
% \textcolor{red}{need to work on the following rest sections}
We introduce semi-stochastic gating parameters that automatically optimize SciML models by applying differentiable neurons on both input and hidden layers, enabling efficient feature selection and model compression.

Each of the $i$-th gating parameter $m^{(l)}_{i}$ at $l$-th layer in the SciML model is associated with a gate, with a value $\pi^{(l)}_{i} \in [0,1]$, which is represented in Equation~(\ref{eq:bern}). During training, $\pi^{(l)}_{i}$ is updated with the gradients of the loss function, dynamically adjusting the neuron’s gating status, along with the optimization of SciML model parameters.
After optimization, if $\pi^{(l)}_{i}$ exceeds the upper threshold $t_u$, the neuron remains active; if it falls below the lower threshold $t_l$, the neuron is deactivated and marked for removal along with the structural consistency tools described in Section~\ref{sec:scd}. For values between $t_l$ and $t_u$, the activation status is determined probabilistically using a Bernoulli distribution parameterized by $\pi^{(l)}_{i}$. 
\begin{equation}
\label{eq:bern}
m^{(i)}_{j} = g(\pi^{(l)}_{i})=
\begin{cases} 
1 & \pi^{(l)}_{i} > t_u, \\
\text{Bern}(\pi^{(l)}_{i}) & t_u \geq \pi^{(l)}_{i} > t_l,\\
0 & t_l \geq \pi^{(l)}_{i}.\\
\end{cases}    
\end{equation}
The approach of employing a threshold to regulate the gate's status is straightforward; however, its discrete nature makes an effective gradient-based optimization infeasible. In Sec~\ref{sec:reparameterization}, we propose a method to approximate the gate in a \textit{differentiable} manner to enhance optimization.

\subsection{Reparametrization for Differentiable Optimization}
\label{sec:reparameterization}

To efficiently optimize gating parameters using gradient-based methods such as Adam~\cite{kingma2014adam} or SGD~\cite{robbins1951stochastic}, we apply a reparameterization trick~\cite{louizos2018learning} that approximates the Bernoulli gate using a hard concrete function. Specifically, for the variable $m^{(i)}{j}$ defined in Equation~(\ref{eq:bern}), we approximate it as follows: 
\begin{equation}
    \label{uniform}
    \begin{aligned}
        & \mnode[fill=teal!15,rounded corners=1pt,inner sep=1pt]{epsilon}{\epsilon}\sim \text{uniform}(t_l,t_u) \\
        & \tilde{m}^{(i)}_{j} = \sigma((\log\frac{\epsilon}{1-\epsilon}+
        \mnode[fill=red!15,rounded corners=1pt,inner sep=1pt]{alpha}{\alpha^{(i)}_{j}})/
        \mnode[fill=green!15,rounded corners=1pt,inner sep=1pt]{tau}{\tau}),
    \end{aligned}
\end{equation}

\begin{tikzpicture}[remember picture,overlay]
\tikzset{
  annot/.style={-{Latex[length=2mm]}, line width=0.35pt, rounded corners=2pt},
  lbl/.style  ={font=\scriptsize, text depth=0, text height=1.2ex, align=center}
}

\path (epsilon.west) coordinate (e_w);
% \coordinate (t_knee) at ($(t_n)+(0,3mm)$);
\coordinate (e_end)  at ($(e_w)+(-10mm,-4mm)$);
\draw[annot, teal!80!black] (e_end) |- (e_w);
\node[lbl, teal!80!black, xshift=0mm, yshift=-1mm, text width=36mm] at (e_end)
  {Randomly sampled\\ seeds};

\path (alpha.south) coordinate (a_s);
\coordinate (a_knee) at ($(a_s)+(0,-4mm)$);
\coordinate (a_end)  at ($(a_s)+(-26mm,-4mm)$);
\draw[annot, red!70!black] (a_end) -| (a_knee) -| (a_s);
\node[lbl, red!70!black, xshift=11mm, yshift=2mm, text width=36mm] at (a_end)
  {Learnable parameters};

\path (tau.north) coordinate (t_n);
% \coordinate (t_knee) at ($(t_n)+(0,3mm)$);
\coordinate (t_end)  at ($(t_n)+(20mm,3mm)$);
\draw[annot, green!80!black] (t_end) -| (t_n);
\node[lbl, green!80!black, xshift=10mm, yshift=8mm, text width=36mm] at (t_n)
  {A temperature controller\\ for smoothness};

\end{tikzpicture}

where $\tau$ is the temperature parameter that controls the smoothness of the relaxation, and $\alpha^{(i)}_{j}$ is a learnable parameter.
This formulation enables gradient propagation through the gating network, allowing simultaneous optimization of both the neuron weights (in the baseline SciML model) and the gate parameters (introduced by \fram).

To further enhance differentiability, we adopt the hard-concrete relaxation~\cite{louizos2017learning}, which stretches the output and clips its value within the range $[0,1]$.
Given the relaxed gating variable $\tilde{m}$ (denoted as $\tilde{m}_j^{(i)}$ for clarity), we have:

\vspace{3mm}
\begin{equation}
    \label{stretch}
    \begin{aligned}
        &\tilde{m}^{(i)}_{j} = \tilde{m}^{(i)}_{j}(\mnode[fill=orange!15,rounded corners=1pt,inner sep=1pt]{zeta}{\zeta} - \gamma) + \mnode[fill=teal!15,rounded corners=1pt,inner sep=1pt]{gamma}{\gamma} \\
        &\tilde{m}^{(i)}_{j} = \mnode[fill=blue!15,rounded corners=1pt,inner sep=1pt]{min}{\min}(1, \mnode[fill=blue!15,rounded corners=1pt,inner sep=1pt]{max}{\max}(0, \tilde{m}^{(i)}_{j})),
    \end{aligned}
\end{equation}

\begin{tikzpicture}[remember picture,overlay]
\tikzset{
  annot/.style={-{Latex[length=2mm]}, line width=0.35pt, rounded corners=2pt},
  lbl/.style  ={font=\scriptsize, text depth=0, text height=1.2ex, align=center}
}

\path (zeta.north) coordinate (zeta_n);
\coordinate (zeta_end)  at ($(zeta_n)+(-20mm,3mm)$);
\draw[annot, orange!80!black] (zeta_end) -| (zeta_n);
\node[lbl, orange!80!black, xshift=-10mm, yshift=5mm, text width=30mm] at (zeta_n)
  {The scaling upper bound};

\path (gamma.north) coordinate (gamma_n);
\coordinate (gamma_end)  at ($(gamma_n)+(20mm,3mm)$);
\draw[annot, teal!80!black] (gamma_end) -| (gamma_n);
\node[lbl, teal!80!black, xshift=10mm, yshift=5mm, text width=36mm] at (gamma_n)
  {The scaling lower bound};

\path (min.south) coordinate (min_s);
\path (max.south) coordinate (max_s);
\coordinate (mm_end)  at ($(min_s)+(-20mm,-8mm)$);
\draw[annot, blue!80!black] (mm_end) -| (min_s);
\draw[annot, blue!80!black] (mm_end) -| (max_s);
\node[lbl, blue!80!black, xshift=10mm, yshift=5mm, text width=36mm] at (mm_end) {Hard-Concrete\\ regularization};

\end{tikzpicture}

\vspace{5mm}

where $\gamma < 0$ and $\zeta > 1$ are hyperparameters that control the scaling and clipping behavior.
%This reparameterization makes the gating process differentiable, enabling the gating parameters and the original neural network parameters to be optimized simultaneously. 

Now we add regularization to the cumulative density $Q$ of the gating variable $\tilde{m}^{(i)}_{j}$, this optimization will lead the gating parameters to become 0 during the training process. The complexity loss, $\mathcal{L}_C$, which measures the SciML model’s flexibility, is expressed as: 

\begin{equation}
\mathcal{L}_C = \sum_{j=1}^{\mnode[fill=yellow!35,rounded corners=1pt,inner sep=1pt]{theta}{|\theta|}} \left( 1 - \mnode[fill=red!15,rounded corners=1pt,inner sep=1pt]{mask}{Q_{\tilde{m}^{(i)}_{j}}}(0 \mid \phi) \right)
\label{lc}
\end{equation}

\begin{tikzpicture}[remember picture,overlay]
\tikzset{
  annot/.style={-{Latex[length=2mm]}, line width=0.35pt, rounded corners=2pt},
  lbl/.style  ={font=\scriptsize, text depth=0, text height=1.2ex, align=center}
}

\path (theta.west) coordinate (theta_w);
\coordinate (theta_end)  at ($(theta_w)+(-20mm,0mm)$);
\draw[annot, yellow!80!black] (theta_end) -- (theta_w);
\node[lbl, yellow!80!black, xshift=-10mm, yshift=5mm, text width=30mm] at (theta_w)
  { \#Gating\\ parameters };

\path (mask.north) coordinate (mask_n);
\coordinate (mask_end)  at ($(mask_n)+(20mm,3mm)$);
\draw[annot, red!80!black] (mask_end) -| (mask_n);
\node[lbl, red!80!black, xshift=12mm, yshift=5mm, text width=36mm] at (mask_n)
  {Cumulative density};%{Final parameter\\ multiplied by weights };

\end{tikzpicture}

The parameters of the distribution are $\phi = (\log\alpha, \tau)$, where $\log\alpha$ is the location and $\tau$ is the temperature.

The complexity loss of the objective in Equation~\ref{lc} under the hard concrete is equivalently expressed as follows:

\vspace{3mm}
\begin{equation}
    \mathcal{L}_C = \sum_{j=1}^{|\theta|} \sigma \left( \mnode[fill=red!15,rounded corners=1pt,inner sep=1pt]{mask}{\log \alpha_j} - \mnode[fill=teal!15,rounded corners=1pt,inner sep=1pt]{beta}{\beta} \log \left( -\frac{\gamma}{\zeta} \right) \right).
    \label{complexity}
\end{equation}

\begin{tikzpicture}[remember picture,overlay]
\tikzset{
  annot/.style={-{Latex[length=2mm]}, line width=0.35pt, rounded corners=2pt},
  lbl/.style  ={font=\scriptsize, text depth=0, text height=1.2ex, align=center}
}

\path (mask.north) coordinate (mask_n);
\coordinate (mask_end)  at ($(mask_n)+(20mm,3mm)$);
\draw[annot, red!80!black] (mask_end) -| (mask_n);
\node[lbl, red!80!black, xshift=12mm, yshift=5mm, text width=36mm] at (mask_n)
  {Learnable gates };%{Final parameter\\ multiplied by weights };

\path (beta.south) coordinate (beta_s);
\coordinate (beta_end)  at ($(beta_s)+(-12mm,-5mm)$);
\draw[annot, teal!70!black] (beta_end) -| (beta_s);
\node[lbl, teal!70!black, xshift=5mm, yshift=2mm, text width=16mm] at (beta_end)
  {$\beta=1/\tau$};

\end{tikzpicture}

% where $|\theta|$ denotes the number of gating parameters added to the model.
In summary, the total optimization is driven by two loss functions: a prediction loss and an additional regularization term $\mathcal{L}_C$ that promotes the (de-)activation of neurons, determining the selection of input features and model parameters ultimately (with the implementation of structure coordination in Section~\ref{sec:scd}).
The combined loss function, $\mathcal{L_T}$, integrates both the accuracy loss ($\mathcal{L}_A$) and the complexity loss ($\mathcal{L}_C$) as follows: 
\begin{equation}
    \mathcal{L_T} = \mathcal{L}_A + \lambda \mathcal{L}_C,
\label{L0_loss_function}
\end{equation}
where $\lambda$ is a hyperparameter that balances $\mathcal{L}_A$ and $\mathcal{L}_C$.

The overall optimization problem is then solved by minimizing the total loss $\mathcal{L_T}$, ensuring that both the model neurons and the distribution of gating parameters converge effectively.  %This approach allows for efficient and adaptive gating parameter, which improves both the sparsity and performance of the model.

\section{Incorporation within SciML Layers}

\label{incorporation}
We integrate gates at the input of each layer so that removing a gate deletes an entire input unit/channel and its fan-in weights. This yields structured sparsity, direct computation, and memory reduction.

\begin{figure}[htb]
    \centering
    \begin{subfigure}[t]{0.47\textwidth}
        % \vspace{5pt}
        \includegraphics[width=0.95\linewidth]{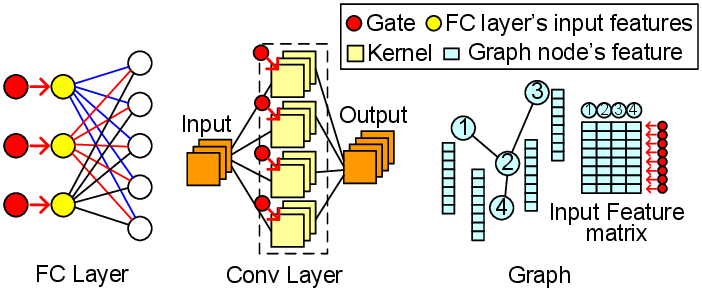}
        % \caption{FLOPs of pre- vs. post-pruning models.}
        \label{flops overall}
    \end{subfigure}
    \hspace{0.05\textwidth}
    \begin{subfigure}[t]{0.47\textwidth}
        \includegraphics[width=0.95\linewidth]{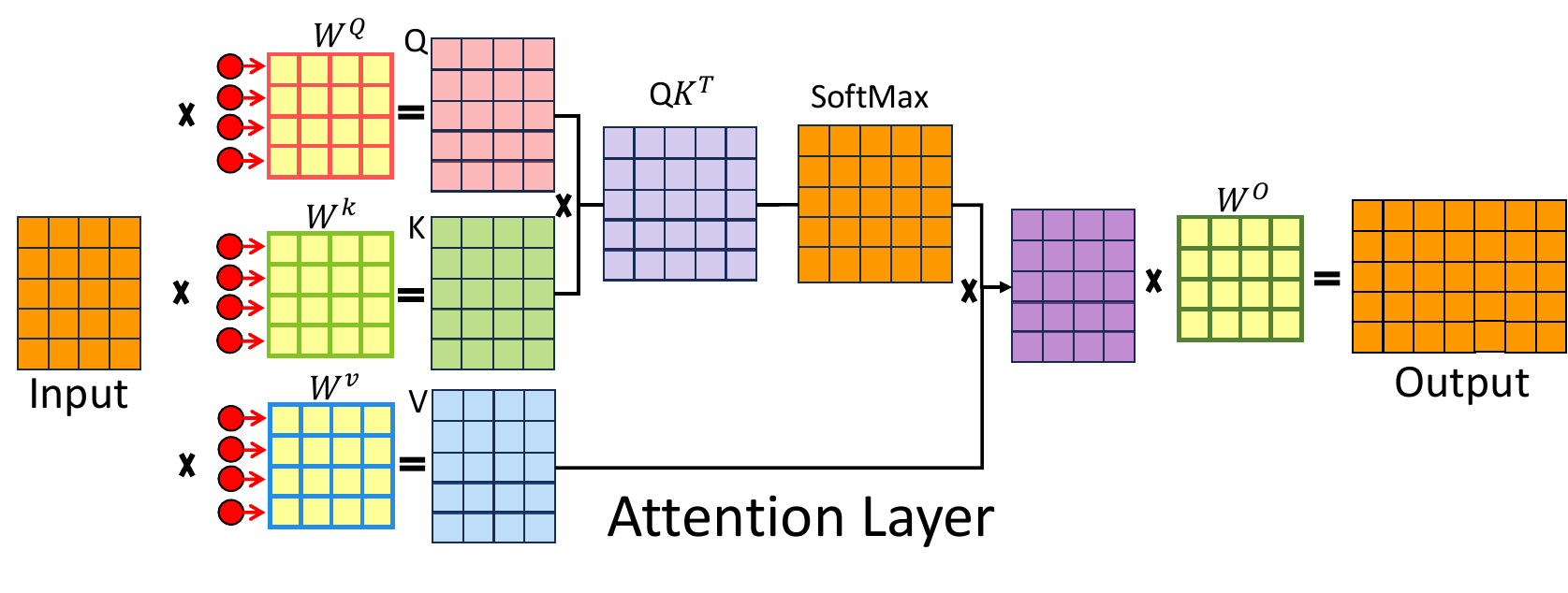}
        % \caption{Peak memory usage of pre- vs. post-pruning models.}
        \label{attention}
    \end{subfigure}
    \caption{Integration with different model layers.}
    \vspace{-3mm}
    \label{fig:draft}
\end{figure}

Specifically, we apply element-wise gating parameters using the element-wise Hadamard product ($\odot$) to regulate the importance of different model components, while preserving the structural consistency of various neural network layers, such as fully connected, convolutional, graph-based, or attention layers (shown in Figure \ref {fig:draft}). 

% We present details in the following text.

\textbf{(1) Fully Connected (FC) Layer}: The FC layer is one of the most widely used types in SciML models for regression problems. Structured pruning is applied to the FC layer, with gating parameters focused on the row structure (input channels).

\begin{equation}
\label{eq:layer}
\mathbf{H}^{(l+1)} = \sigma\left(\mnode[fill=yellow!35,rounded corners=1pt,inner sep=1pt]{prop}{\mathbf{X}^{(l)}} \cdot \mnode[fill=teal!15,rounded corners=1pt,inner sep=1pt]{eps}{\mathbf{m}^{(l)}} \odot \mnode[fill=orange!15,rounded corners=1pt,inner sep=1pt]{emb}{\mathbf{W}^{(l)}} + \mathbf{b}^{(l)}\right).
\end{equation}

\begin{tikzpicture}[remember picture,overlay]
\tikzset{
  annot/.style={-{Latex[length=2mm]}, line width=0.35pt, rounded corners=2pt},
  lbl/.style  ={font=\scriptsize, text depth=0, text height=1.2ex, align=center}
}

\path (eps.north) coordinate (eps_n);
\coordinate (eps_end)  at ($(eps_n)+(26mm,3mm)$);
\draw[annot, teal!80!black] (eps_end) -| (eps_n);
\node[lbl, teal!80!black, xshift=12mm, yshift=5mm, text width=36mm] at (eps_n)
  {Gating parameters ($\mathbb{R}^{1 \times n}$)};

\path (prop.south) coordinate (prop_s);
\coordinate (prop_end)  at ($(prop_s)+(-22mm,-7mm)$);
\draw[annot, yellow!80!black] (prop_end) -| (prop_s);
\node[lbl, yellow!80!black, xshift=10mm, yshift=5mm, text width=36mm] at (prop_end) {Input feature\\ $\mathbb{R}^{m \times n}$};

\path (emb.south) coordinate (emb_s);
\coordinate (emb_end)  at ($(emb_s)+(20mm,-7mm)$);
\draw[annot, orange!80!black] (emb_end) -| (emb_s);
\node[lbl, orange!80!black, xshift=-10mm, yshift=5mm, text width=36mm] at (emb_end) {Layer weights\\ $\mathbb{R}^{n \times k}$};

\end{tikzpicture}

FC layers are ubiquitous in SciML models and form foundational components of CNNs, GINs, GCNs, and Transformers. We set the gating vector to match the number of rows (input units) in each FC layer, so that each gate modulates a group of FC weights. This design enables structured pruning in a principled, unit- or group-wise manner.

\textbf{(2) Convolutional (CONV) Layer:}
The gating variables are integrated into the convolutional kernels depicted in yellow in Figure~\ref{fig:draft}. The gating parameter $\mathbf{m}^{(l)} \in \mathbb{R}^{1\times d}$ is applied to the output channels:

\begin{equation}
    \mathbf{H}^{(l+1)} = \sigma\left(\mnode[fill=teal!15,rounded corners=1pt,inner sep=1pt]{conv}{\text{conv}}(\mathbf{X}^{(l)}, \mathbf{m}^{(l)} \odot \mathbf{W}^{(l)}) + \mathbf{b}^{(l)}\right)
\end{equation}

\begin{tikzpicture}[remember picture,overlay]
\tikzset{
  annot/.style={-{Latex[length=2mm]}, line width=0.35pt, rounded corners=2pt},
  lbl/.style  ={font=\scriptsize, text depth=0, text height=1.2ex, align=center}
}

\path (conv.north) coordinate (conv_n);
\coordinate (conv_end)  at ($(conv_n)+(22mm,3mm)$);
\draw[annot, teal!80!black] (conv_end) -| (conv_n);
\node[lbl, teal!80!black, xshift=10mm, yshift=5mm, text width=36mm] at (conv_n)
  {Convolution operation};
  
\end{tikzpicture}

% where:
% \begin{itemize}
%     \item $\mathbf{X}^{(l)} \in \mathbb{R}^{b \times c \times h \times w}$ represents the input feature maps with batch size $b$, number of channels $c$, height $h$, and width $w$.
%     \item $\mathbf{W}^{(l)} \in \mathbb{R}^{c \times d \times k \times k}$ denotes the convolutional kernels, where $d$ is the number of output channels, $c$ is the number of input channels, and $k$ is the kernel size.
% \end{itemize}

% \textcolor{blue}{Pruning an input channel needs to remove the corresponding fan-out neurons across \emph{all} output filters; following FC layers consuming flattened features are updated via our consistency mapper (Sec.~\ref{sec:scd}).}

\textbf{(3) Graph Isomorphism Network (GIN) Layer:}
There are three stages in the GIN layer: embedding edge feature, aggregating neighboring nodes, and linearly
map aggregated nodes. In the GIN layer, we applied the same gating parameters $\mathbf{m}^{(l)} \in \mathbb{R}^{1\times n}$ to both the output of aggregation nodes and the input nodes to maintain the structure pruning consistency:

\vspace{3mm}
\begin{equation}
    \begin{aligned}
        &\mathbf{H}^{(l+1)} = \sigma(\mathbf{W}^{(l)}((1 + \mnode[fill=teal!15,rounded corners=1pt,inner sep=1pt]{eps}{\varepsilon}) \cdot \mathbf{m}^{(l)}\odot \mathbf{X}^{(l)} + \\
        &\mathbf{m}^{(l)} \odot \mnode[fill=blue!15,rounded corners=1pt,inner sep=1pt]{prop}{prop}(\mnode[fill=orange!15,rounded corners=1pt,inner sep=1pt]{emb}{emb}(\mathbf{E}), \mathbf{X}^{(l)})))
    \end{aligned}
\end{equation}

\begin{tikzpicture}[remember picture,overlay]
\tikzset{
  annot/.style={-{Latex[length=2mm]}, line width=0.35pt, rounded corners=2pt},
  lbl/.style  ={font=\scriptsize, text depth=0, text height=1.2ex, align=center}
}

\path (eps.north) coordinate (eps_n);
\coordinate (eps_end)  at ($(eps_n)+(20mm,3mm)$);
\draw[annot, teal!80!black] (eps_end) -| (eps_n);
\node[lbl, teal!80!black, xshift=10mm, yshift=5mm, text width=36mm] at (eps_n)
  {Learnable parameter};

\path (prop.south) coordinate (prop_s);
\coordinate (prop_end)  at ($(prop_s)+(-22mm,-7mm)$);
\draw[annot, blue!80!black] (prop_end) -| (prop_s);
\node[lbl, blue!80!black, xshift=10mm, yshift=5mm, text width=36mm] at (prop_end) {Aggregates neighbor\\ information};

\path (emb.south) coordinate (emb_s);
\coordinate (emb_end)  at ($(emb_s)+(25mm,-4mm)$);
\draw[annot, orange!80!black] (emb_end) -| (emb_s);
\node[lbl, orange!80!black, xshift=-10mm, yshift=2mm, text width=36mm] at (emb_end) {Embeds edge attributes};

\end{tikzpicture}

\vspace{5mm}

where:
\begin{itemize}
    \item $\mathbf{E} \in \mathbb{R}^{1 \times e}$: Edge features, embedded to $\mathbb{R}^{e \times n}$.
    % \item $\mathbf{X}^{(l)} \in \mathbb{R}^{m \times n}$: Node features ($m$ nodes, $n$-dimensional).
    % \item $\mathbf{H}' \in \mathbb{R}^{m \times n}$: Aggregated output.
    % \item $\mathbf{W}^{(l)} \in \mathbb{R}^{n \times d}$: Weight matrix ($n$ input, $d$ output channels).
\end{itemize}

\textbf{(4) Graph Convolutional Network (GCN) Layer:} A GCN layer~\cite{kipf2016semi} defines a first-order approximation of a localized spectral filter on graphs. GCNs can be understood as a generalization of convolutional neural networks to graph-structured data.
% For the GCN layer, we applied the same gating parameter $\mathbf{m}^{(l)} \in \mathbb{R}^n$ to the input channels of the weights $\mathbf{W}^{(l)}_1$ and $\mathbf{W}^{(l)}_2$ to maintain the structure pruning character of \fram:
To keep structural consistency, we also apply the same gating parameters $\mathbf{m}^{(l)} \in \mathbb{R}^{1\times n}$ to the input channels $\mathbf{W}^{(l)}_1$ and $\mathbf{W}^{(l)}_2$:

\begin{equation}
\vspace{4mm}
    \begin{aligned}
        &\mathbf{H}' = \mathbf{W}^{(l)}_1\cdot\mathbf{X}^{(l)}+b^{(l)}_1\\
        &\mathbf{E}' = \mathbf{W}^{(l)}_2\cdot \mathbf{E}+b^{(l)}_2\\
        \mathbf{H}^{(l)} = \mathbf{m}^{(l)} \odot &prop(\mnode[fill=yellow!35,rounded corners=1pt,inner sep=1pt]{h}{\mathbf{H}'}, \mnode[fill=orange!15,rounded corners=1pt,inner sep=1pt]{e}{\mathbf{E}'}) + \mathbf{m}^{(l)} \odot \text{ReLU}(\mathbf{H}' + emb(\mnode[fill=teal!15,rounded corners=1pt,inner sep=1pt]{r}{\mathbf{R}}))
    \end{aligned}
\end{equation}
\begin{tikzpicture}[remember picture,overlay]
\tikzset{
  annot/.style={-{Latex[length=2mm]}, line width=0.35pt, rounded corners=2pt},
  lbl/.style  ={font=\scriptsize, text depth=0, text height=1.2ex, align=center}
}

\path (r.north) coordinate (r_n);
\coordinate (r_end)  at ($(r_n)+(-15mm,3mm)$);
\draw[annot, teal!80!black] (r_end) -| (r_n);
\node[lbl, teal!80!black, xshift=6mm, yshift=2mm, text width=36mm] at (r_end)
  {Root node};

\path (h.south) coordinate (h_s);
\coordinate (h_end)  at ($(h_s)+(-22mm,-7mm)$);
\draw[annot, yellow!80!black] (h_end) -| (h_s);
\node[lbl, yellow!80!black, xshift=10mm, yshift=5mm, text width=36mm] at (h_end) {Transformed node\\ features $\mathbb{R}^{m \times d}$};

\path (e.south) coordinate (e_s);
\coordinate (e_end)  at ($(e_s)+(22mm,-7mm)$);
\draw[annot, orange!80!black] (e_end) -| (e_s);
\node[lbl, orange!80!black, xshift=-10mm, yshift=5mm, text width=36mm] at (e_end) {Transformed edge\\ features $\mathbb{R}^{e \times d}$};

\end{tikzpicture}

\textbf{(5) Attention Layer:}
For attention layers, gating parameters are applied to three input weight matrices: $\mathbf{W}_Q$, $\mathbf{W}_K$, and $\mathbf{W}_V$. All gating parameters are applied to the first dimension of weight matrices, as illustrated in Figure~\ref{fig:draft} and equations~\ref{attention}.

\begin{equation}
    \begin{aligned}
        \mathbf{Q}^{(l)} &= \mathbf{X}^{(l)} \cdot \mathbf{m}^{(l)}_Q \odot \mathbf{W}^{(l)}_Q \\
        \mathbf{K}^{(l)} &= \mathbf{X}^{(l)} \cdot \mathbf{m}^{(l)}_K \odot \mathbf{W}^{(l)}_K \\
        \mathbf{V}^{(l)} &= \mathbf{X}^{(l)} \cdot \mathbf{m}^{(l)}_V \odot \mathbf{W}^{(l)}_V \\
        \mathbf{H}^{(l+1)} &= \mnode[fill=orange!15,rounded corners=1pt,inner sep=1pt]{att}{\text{Attention}}(\mathbf{Q}^{(l)}, \mathbf{K}^{(l)}, \mathbf{V}^{(l)})  \mathbf{W}^{(l)}_O
    \end{aligned}
    \label{attention}
    \vspace{4mm}
\end{equation}
\begin{tikzpicture}[remember picture,overlay]
\tikzset{
  annot/.style={-{Latex[length=2mm]}, line width=0.35pt, rounded corners=2pt},
  lbl/.style  ={font=\scriptsize, text depth=0, text height=1.2ex, align=center}
}

\path (att.south) coordinate (att_s);
\coordinate (att_end)  at ($(att_s)+(22mm,-5mm)$);
\draw[annot, orange!80!black] (att_end) -| (att_s);
\node[lbl, orange!80!black, xshift=-10mm, yshift=3mm, text width=36mm] at (att_end) {$\text{softmax}\left(\frac{\mathbf{Q}\mathbf{K}^T}{\sqrt{d}}\right)\mathbf{V}$};
\end{tikzpicture}

This gating mechanism can be seamlessly integrated into multi-head attention, enabling structured pruning while preserving the overall attention architecture.

\section{Structural Consistency}
\label{sec:scd}

% To enable seamless extraction and deployment of pruned SciML models, we introduce a structural consistency mechanism that ensures correctness, compatibility, and computational efficiency post-pruning. Specifically, \fram constructs a layer-wise mapping system that tracks preserved input and output indices, enabling accurate feature alignment across layers. Additionally, we design specialized coordination strategies to handle complex architectural transformations—such as transitions between convolutional and fully connected layers, or concatenation operations—ensuring structural integrity throughout.

After the gate indication for neuron selection, the corresponding removal must maintain the remaining network executable: tensor dimensions must align, layer dependencies must hold, and model outputs must stay numerically valid. To guarantee these, we introduce a structural consistency mechanism that performs two core functions:
(i) it tracks which channels, neurons, or features are preserved across layers, and
(ii) it reconciles architectural transformations such as flattening, concatenation, or embedding, so that input–output mappings remain consistent.

\subsection{Layer-Wise Consistency Mapping}

\fram maintains a consistency dictionary for each layer, composed of [$input\_mask$, $output\_mask$] pairs. These masks encode the indices of preserved weights and propagate remove decisions across layers.
The $input\_mask$ represents the indices of the input channels or neurons that are retained (those with gate values equal to ones). The $output\_mask$ represents the indices of the output channels.
Below we present two representative mapping scenarios in \fram.

\subsubsection{Between the CONV and FC layer}
\begin{figure}[!ht]
    \centering
    \includegraphics[width=0.48\textwidth]{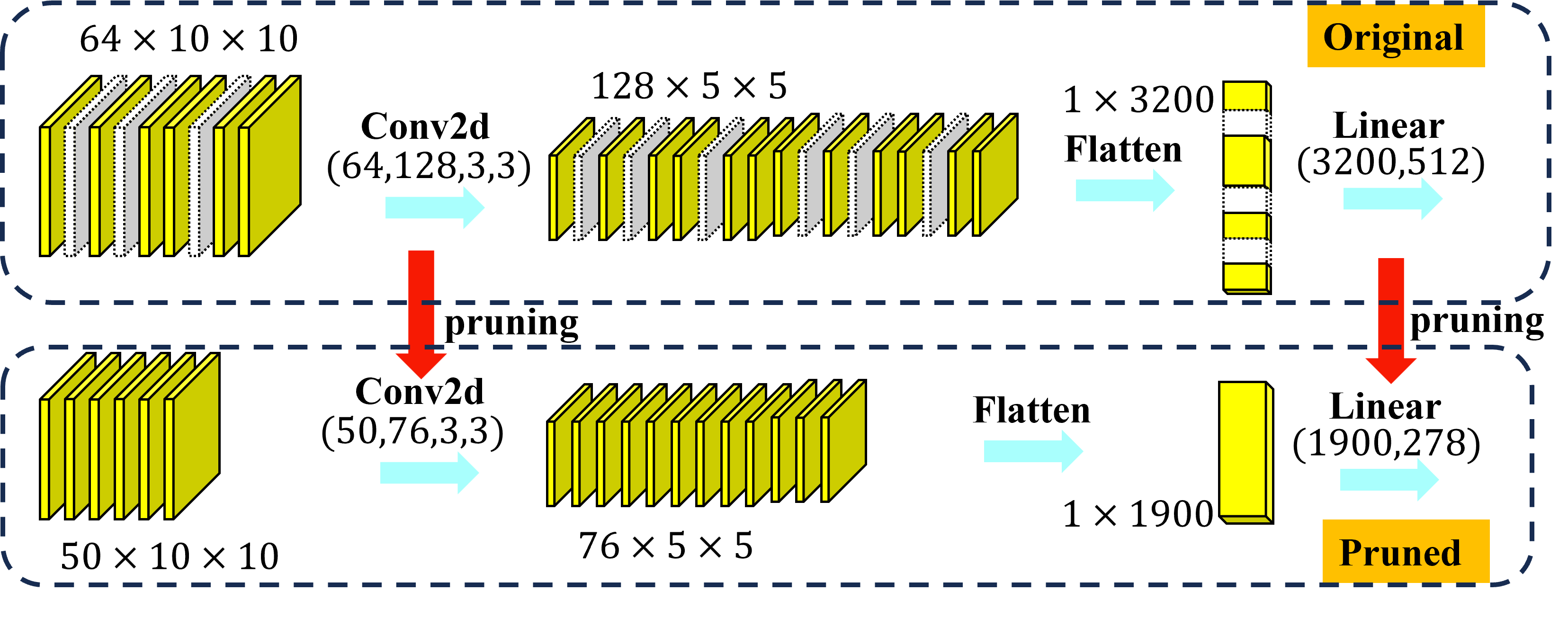}
    \caption{Structured transition between the CONV and FC layer. When a CONV layer is followed by an FC layer, the output of the CONV layer serves as the input to the FC layer; therefore, the redundant outputs corresponding to the deactivated neurons of the CONV layer must also be removed to maintain tensor alignments.}
    \label{fig:image1}
\end{figure}

In the transition from a CONV layer to an FC layer, the output of the CONV layer with spatial dimensions is flattened to serve as the input to the FC layer. We implement a function \texttt{CONV2FC\_dictionary()} in \fram for index-mapping. 
For example, as shown in Fig.~\ref{fig:image1}, if the CONV layer contains weights $\mathbf{W} \in \mathbb{R}^{64 \times 128 \times 3 \times 3}$ and a pruning operation reduces the input channels from 64 to 50, the mapper adjusts the $output\_mask$ accordingly.\footnote{The translation follows the equation of $\text{input\_mask}[i] = \text{output\_mask}[j] \times w \times h + k$, where $w$ and $h$ are the width and height of the output feature map from the CONV layer, and $k$ is the specific position within each spatial dimension of the output feature map, ranging from $0$ to $(w \times h - 1)$.}

\subsubsection{Between the embedding and FC layer}
% In the transition from an embedding layer to an FC layer, the output of the embedding layer is used to serve as the input to the FC layer. In practice, we only apply gating parameters to the FC layer while not applying gating parameters to the embedding layer. Hence, in order to maintain consistency and structure pruning, the embedding layer and FC layer share the same gating parameters, which means the $output\_mask$ of the node and edge embeddings equals the $input\_mask$ of the FC layer. As shown in Fig.~\ref{fig:embedding}, we use a graph with five nodes and four edges as an input example for the embedding layer, and each node and edge is encoded into a 300-dimensional vector. After \fram deactivate the subsequent FC layers, the input weights are reduced to 189 for node embeddings and 130 for edge embeddings. Consequently, the weights of the node and edge embeddings will be reduced to 189 and 130, respectively.
In the transition from an embedding layer to a fully connected (FC) layer, the output vectors from the embedding layer serve as the input to the FC layer.
In our design, gating parameters are applied only to the FC layer rather than directly to the embedding layer.
To maintain structural consistency between these two layers, the embedding and FC layers share the same gating configuration.
This means that when certain input channels of the FC layer are deactivated, the corresponding output dimensions of the embedding layer are also pruned to ensure proper tensor alignment.

\begin{figure}[!ht]
    \centering
    \includegraphics[width=0.48\textwidth]{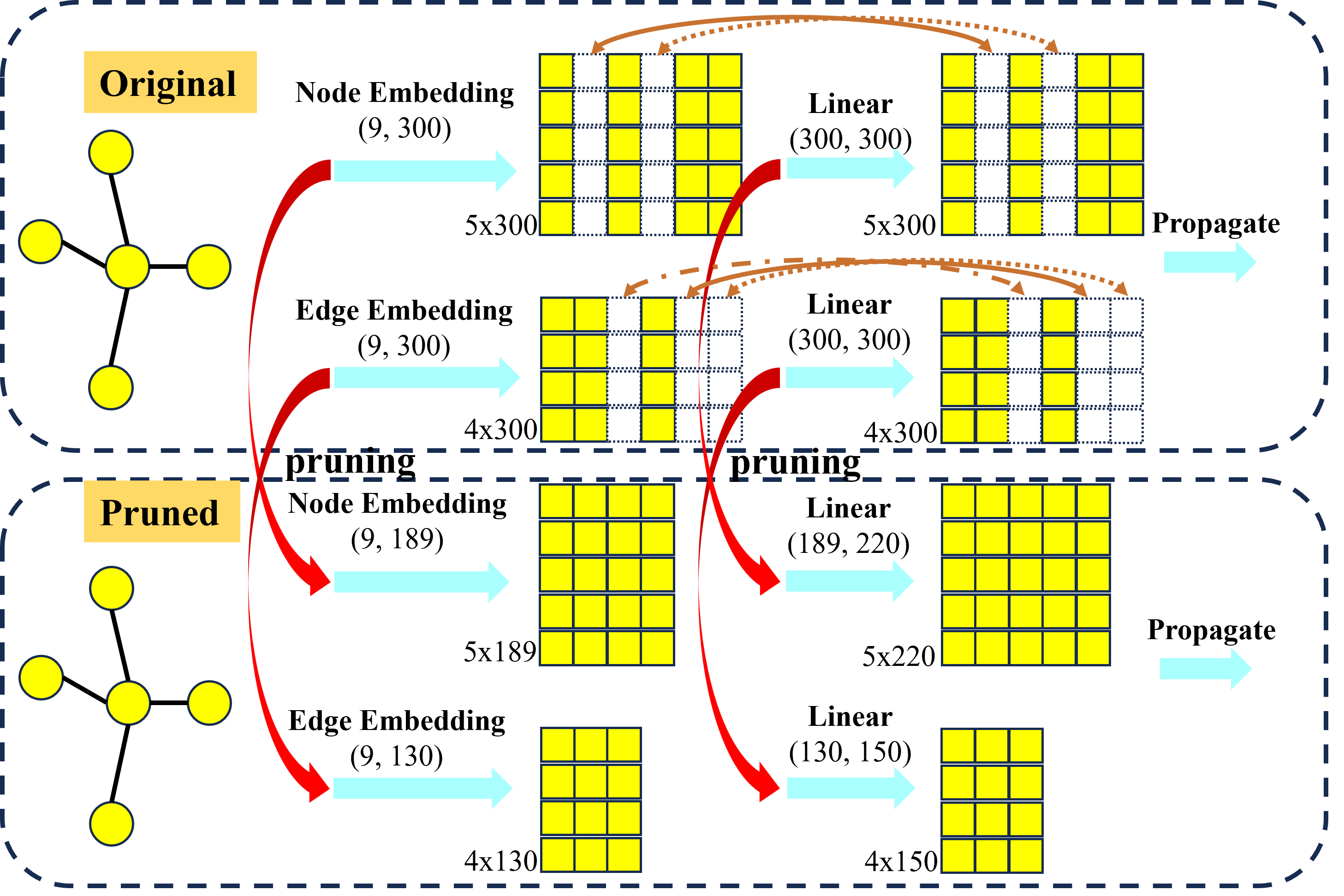}
    \caption{Structured transition between the embedding and FC layer. When input features are deactivated, the corresponding weights in the following FC layer must also be removed.}
    \label{fig:embedding}
\end{figure}

As shown in Figure~\ref{fig:embedding}, we take a graph with five nodes and four edges as an example.
Each node and edge is initially embedded into a 300-dimensional feature vector, resulting in a node embedding matrix of size $5\times300$ and an edge embedding matrix of size $4\times300$.
After \fram optimizes the downstream FC layers, the gating mechanism identifies and removes redundant input channels, reducing the effective embedding dimensions to 189 for node features and 130 for edge features.
Consequently, the corresponding FC layers are updated to accept inputs of sizes $189$ and $130$, respectively, maintaining structural alignment between the embedding and FC layers after optimization.

\subsection{Cross-Layer Topology Coordinator}
When layers are concatenated, reshaped, or otherwise interconnected, maintaining consistent index alignment across them becomes non-trivial.
The \textit{Topology Coordinator} in \fram addresses this challenge by systematically managing index mappings to ensure that optimization decisions, such as channel or neuron deactivation, are correctly propagated across dependent layers.
This mechanism prevents tensor misalignment, preserves architectural integrity, and enables end-to-end structural optimization without user intervention.

\begin{figure}[!ht]
    \centering
    % \vspace{-10pt}
    \includegraphics[width=0.49\textwidth]{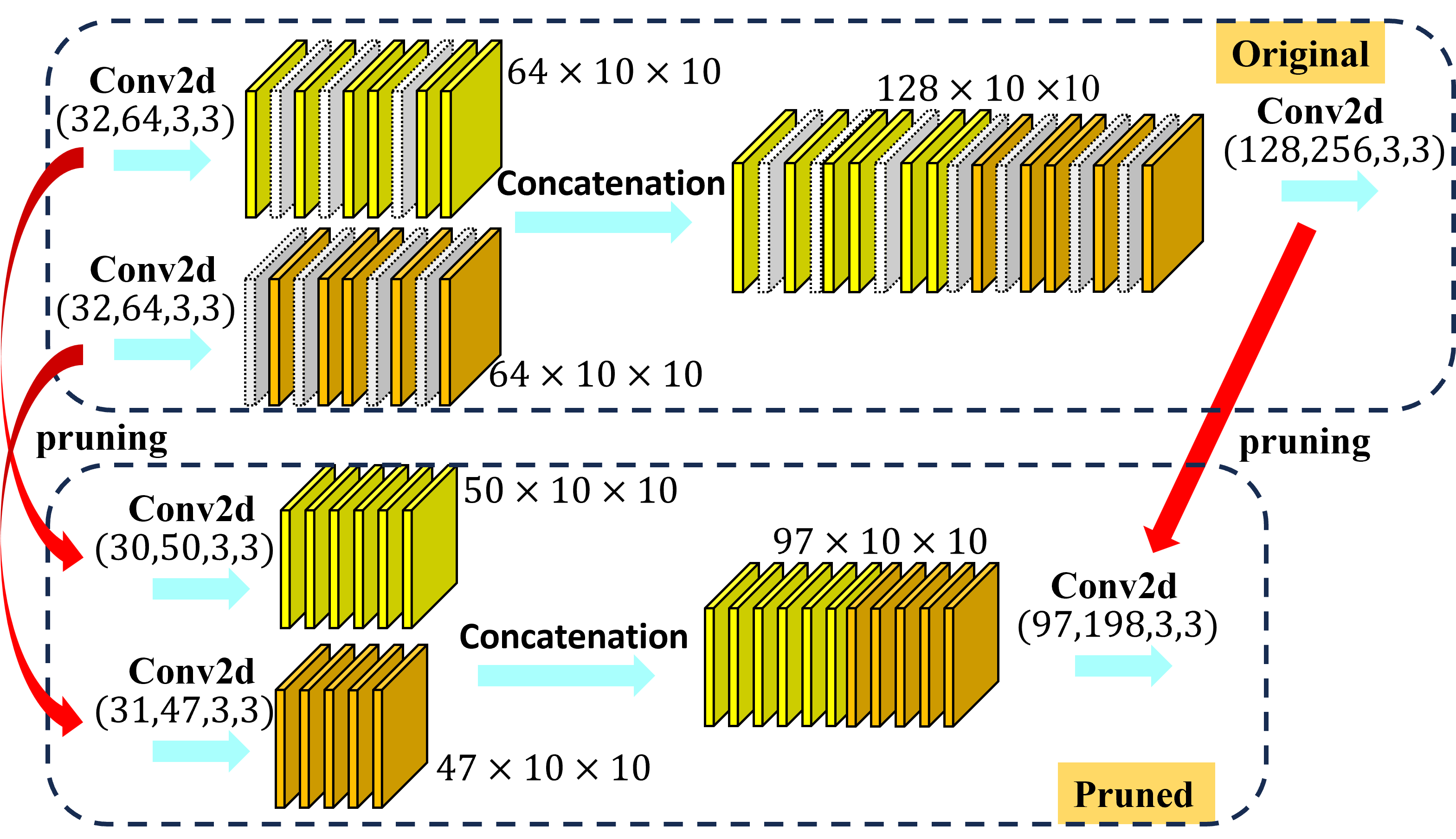}
    \caption{Dependency between concatenation. When two outputs of two CONV layers are concatenated before being connected to another CONV layer, the indices of the CONV kernels must be remapped to maintain structural integrity.}
    \label{fig:image2}
    % \vspace{-1em}
\end{figure}

Consider the case of concatenating two convolutional layers after optimization.
As with flattening operations, it is essential to align the preserved indices in the respective dictionaries <$out_mask_1$, $out_mask_2$> with the $in_mask$ of the downstream concatenated layer.
In the example shown in Figure~\ref{fig:image2}, the first convolutional layer reduces its output channels from 64 to 50, and the second from 64 to 47.
After concatenation, the combined input to the subsequent layer consists of 97 channels instead of 128.
The Topology Coordinator automatically merges the retained indices from both layers, applying the necessary offset to the second set to maintain correct ordering and ensure seamless connectivity

Extracting consistent subnetworks from heterogeneous neural architectures is inherently complex.
\fram streamlines this process by encapsulating all required functionalities, including index mapping, mask propagation, and model extraction.
Users only need to specify their model and dataset; \fram autonomously manages the structural adjustments, delivering a compact, deployable model that preserves both functionality and layer connectivity.

\begin{figure}[!ht]
    \centering
    % \vspace{-10pt}
    \includegraphics[width=0.45\textwidth]{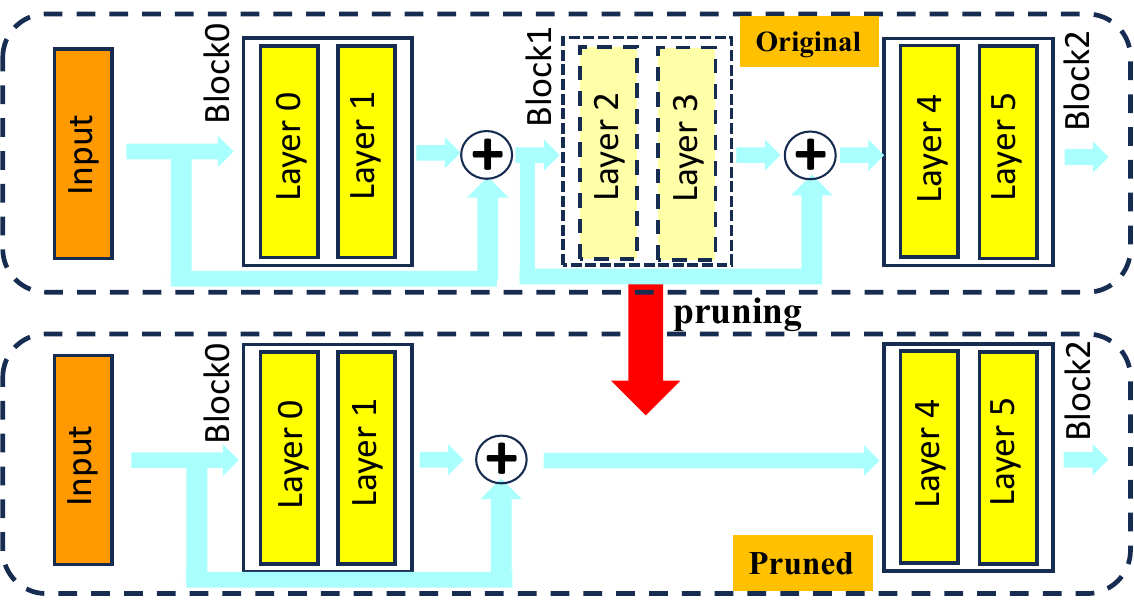}
    \caption{%Prune layer within residual architecture. \fram can skip the layer if all the elements within the block are selected as non-active.
    Layer-wise pruning within a residual architecture.}
    \label{fig:rmlayer}
    % \vspace{-1em}
\end{figure}

\subsection{Residual-Aware Layer Removal}
\fram can also remove redundant layers from a given model architecture. Figure~\ref{fig:rmlayer} illustrates the pruning process applied inside a residual block. In the original structure (top), the residual branch contains an intermediate transformation layer (highlighted by the dashed box). During pruning, if all parameters of \textit{Layer2} or \textit{Layer3} are pruned by \fram, then \textit{Block1} is identified as redundant and needs to be removed. Consequently, \fram will directly connect the remaining blocks of the residual branch to the addition operation, while preserving compatibility with the skip connection.
%This operation reduces both parameter count and computational cost, yet maintains computation consistency so that residual addition remains valid. The process highlights how \fram can simplify residual architectures by eliminating non-essential components while retaining functional equivalence.

\section{Performance Evaluation}
\label{sec:evaluation}

\noindent \textbf{Platform.}
All experiments are conducted on a computing cluster equipped with eight NVIDIA A100 GPUs, each with 80 GB of memory. The cluster is also powered by 96 Intel Xeon Gold 6342 CPU cores, operating at 2.80 GHz. The software environment includes CUDA version 12.3 and NVIDIA Driver version 545.23.08. Model training and inference are performed using Python 3.11 and PyTorch 2.2.

\noindent \textbf{Workloads.}
Table \ref{datasets} presents the diverse scientific workloads evaluated in this study. We assess the performance of \fram across various applications, including fluid dynamics (CFD, Fluid), molecular dynamics (PureMD), cosmology (CosmoFlow), material science (DMS, EM-denoise, StemDL), physics (Optical, SLSTR), biology (PPA), chemistry (Molhiv), and medical science (Brain Tumor). These workloads feature a range of input modalities such as images, vectors, matrices, and graphs, as well as target tasks like regression and classification. To better reflect the multi-sample, multi-iteration nature of SciML inference in real applications, we adopt \textit{dynamic batch size configurations}\footnote{A state-of-the-art (SOTA) evaluation strategy widely used to assess AI model efficiency, as demonstrated in works like ORCA~\cite{280922}.}, with an upper bound of GPU memory usage for fair comparison.

\begin{table}[!ht]
\centering
\caption{Summary of evaluated scientific workloads}
\label{datasets}
\begin{adjustbox}{width=0.5\textwidth,center}
{\tiny
\setlength{\tabcolsep}{1pt}
\renewcommand{\arraystretch}{1.2}
\begin{tabular}{@{}p{1.4cm}p{1.7cm}p{1.3cm}p{3.5cm}@{}}
\toprule
\textbf{Workloads} & \textbf{Domain} & \textbf{Network} & \textbf{Task} \\
\midrule
CFD~\cite{dong2023auto}                       & Fluid Simulation            & FCs                             & Surrogate:\texttt{Compute\_Flux()}\\
Fluid~\cite{dong2023auto}                     & Fluid Simulation            & FCs                             & Surrogate:\texttt{Compute\_Force()} \\
Puremd\cite{dong2023auto}                     & Molecular                   & FCs                             & Surrogate:\texttt{Add\_dBond\_to\_Forces()}   \\
CosmFlow\cite{mathuriya2018cosmoflow}         & Cosmology                   & CONVs+FCs                       & Regression: The universe parameters \\
EM-DN\cite{thiyagalingam2022scientific}       & Material Sciences           & Unet                            & Regression: Denoising     \\
Synthetic\cite{thiyagalingam2022scientific}   & Math formulation            & FCs                             & Regression: Math formulation     \\
Optical\cite{thiyagalingam2022scientific}     & Instrumentation             & Enc.\&Dec.                      & Detection: Optical equipment damage \\
DMS\cite{thiyagalingam2022scientific}         & Material Sciences           & CONVs+FCs                       & Classification: Image  \\
% MNIST~\cite{lecun1998gradient}                & Computer vision             & FCs                             & Classification: Image \\
% Cifar10~\cite{krizhevsky2009learning}         & Computer vision             & AlexNet                         & Classification: Image \\
STEM-DL\cite{thiyagalingam2022scientific}     & Material Sciences           & VGG                             & Classification: Diffraction patterns \\
SLSTR\cite{thiyagalingam2022scientific}       & Atmospheric                 & Unet                            & Classification: Image(at pixel level) \\
PPA~\cite{szklarczyk2019string}               & Biological Science          & GCN                             & Classification: Taxonomic group \\
Molhiv~\cite{wu2018moleculenet}               & Chemical Science            & GIN                             & Classification: Molecular properties\\
BrainTumor~\cite{kaynaaf_brain_tumour_mri}    & Medical                     & ViT                             & Classification: Tumor detection\\
\bottomrule
\vspace{-2em}
\end{tabular}}
\end{adjustbox}
\end{table}

\begin{figure*}[!ht]
    \centering
    % 第一行
        \begin{subfigure}[b]{0.49\textwidth}
        \includegraphics[width=\textwidth]{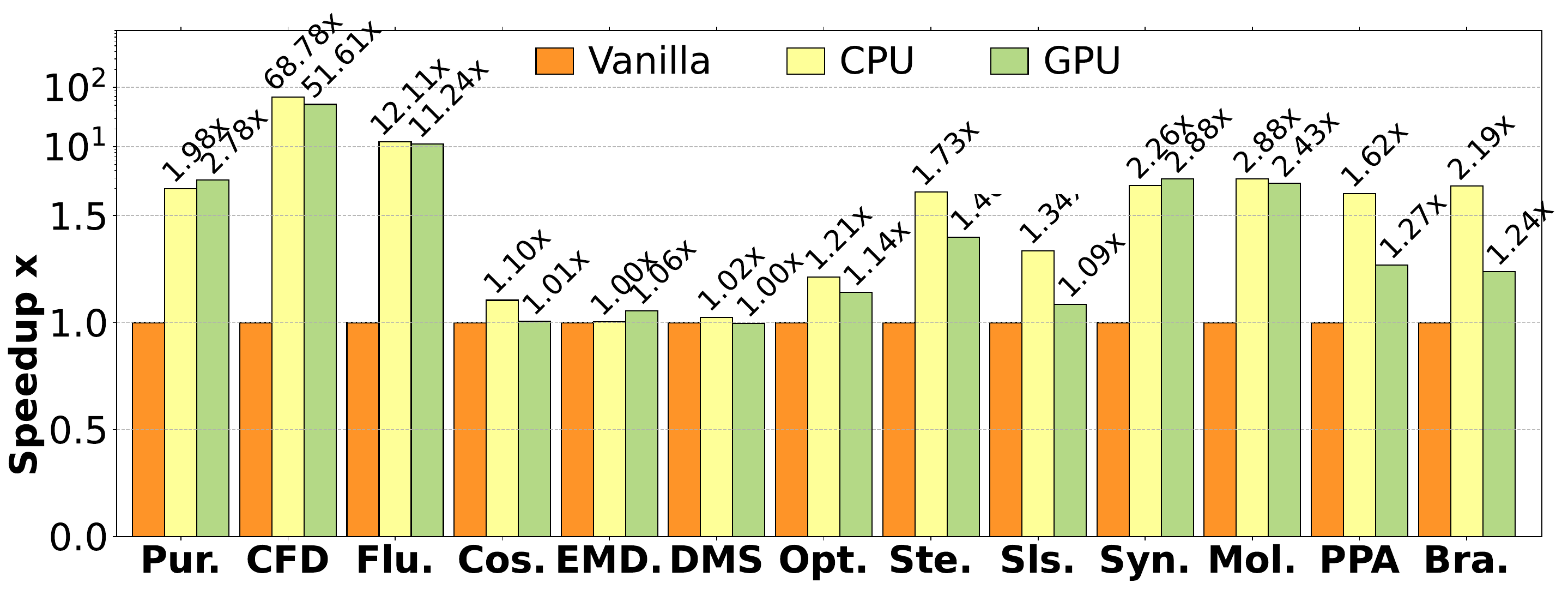}
        \caption{Speedup}
        \label{fig:speedup}
    \end{subfigure}
    \hfill
    \begin{subfigure}[b]{0.49\textwidth}
        \includegraphics[width=\textwidth, height=0.14\textheight]{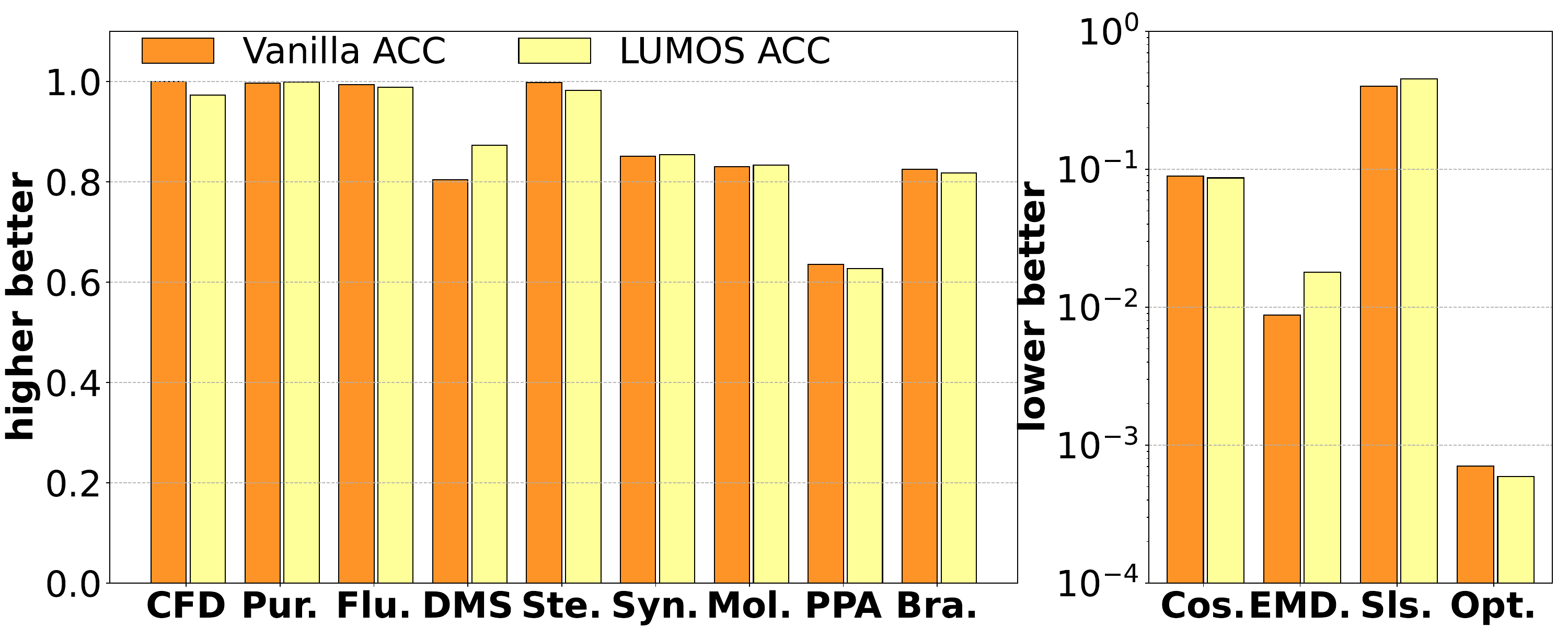}
        \caption{Accuracy}
        \label{Accuracy overall}
    \end{subfigure}
         \vspace{0.5em} % 控制上下间距
         
    % 第二行
    \begin{subfigure}[b]{0.49\textwidth}
        \includegraphics[width=\textwidth]{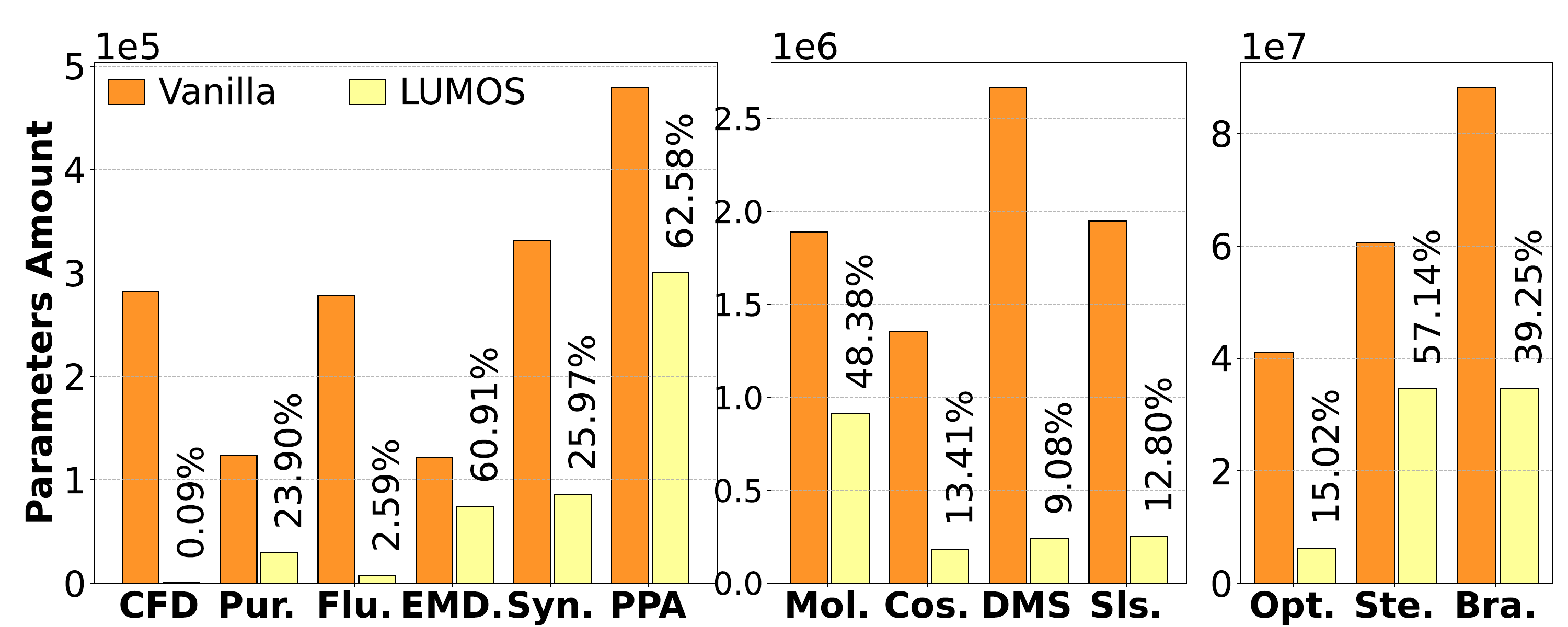}
        \caption{Parameters}
        \label{fig:parameter_reduction}
    \end{subfigure}
    \hfill
    \begin{subfigure}[b]{0.49\textwidth}
        \includegraphics[width=\textwidth]{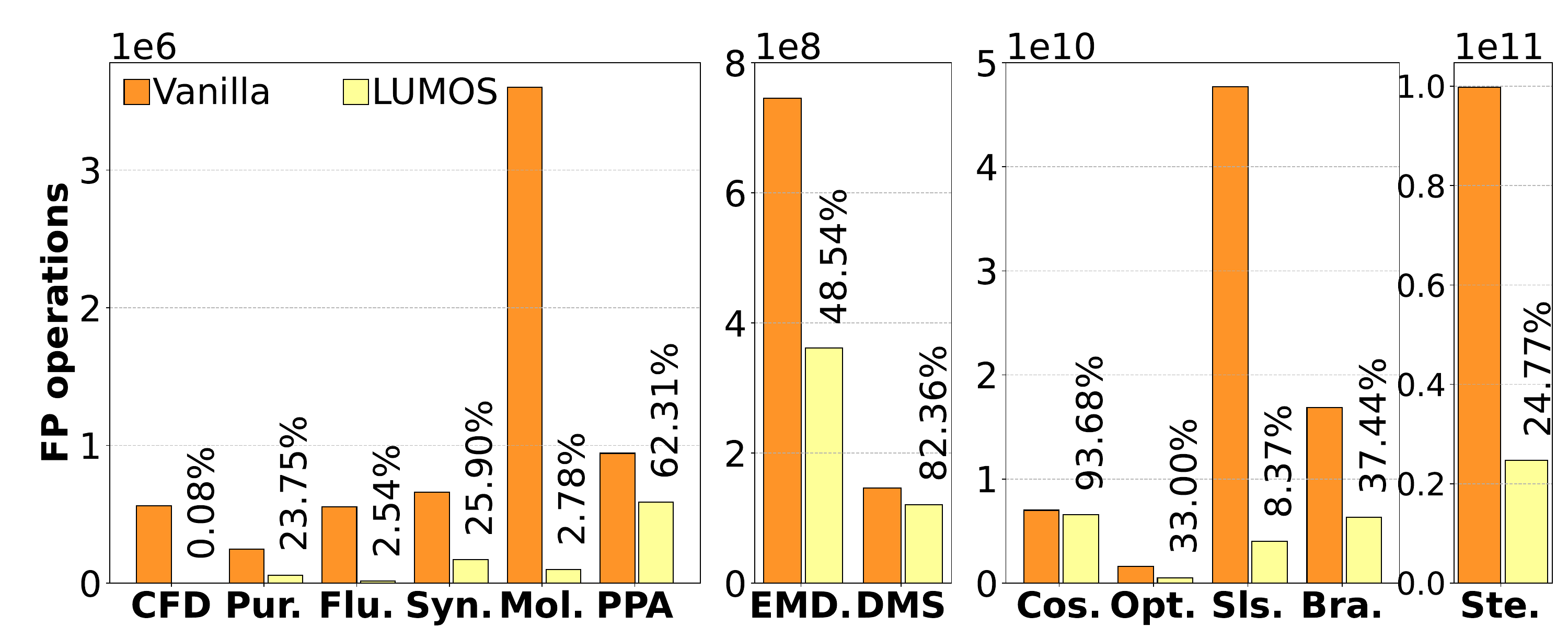}
        \caption{Floating point operations}
        \label{fig:flops}
    \end{subfigure} 

    \caption{Comparison between baseline SciML models and \fram across latency, accuracy, parameter count, and FLOPs.}
    \label{fig:overall}
    \vspace{-3mm}
\end{figure*}

\subsection{Overall Performance}
We evaluate the overall performance of \fram with four metrics: speedup (time-to-solution), accuracy, parameter reduction, and floating point operations (FLOPs). Each reflects a unique angle of \fram's efficiency. 

\ding{182} \textit{Speedup} quantifies the reduction in inference time after applying \fram.  Figure~\ref{fig:speedup} shows the end-to-end (E2E) latency improvements on two major hardware platforms: the NVIDIA A100 GPUs and Intel Xeon Gold 6342 CPUs.
On GPUs, \fram achieves up to a $51.6\times$ speedup (CFD), with an average acceleration of $6.4\times$ across all evaluated workloads. On CPUs, the speedup reaches $68.8\times$ (CFD), averaging $7.6\times$ overall.
SciML workloads, particularly surrogate models, benefit significantly since their computational demands are linear with respect to model parameters. Even within architectures primarily featuring convolutional or graph operations, \fram demonstrates measurable speedups while maintaining accuracy.

\ding{183} \textit{Accuracy} assesses the predictive quality of the optimized SciML models. We employ three standard metrics corresponding to the type of scientific task: \textbf{Acc}, $R^2$ score, and \texttt{MSE}.
For classification-based SciML tasks, \textbf{Acc} measures the ratio of correctly predicted instances to the total number of samples, which is applied to the workloads of DMS, STEM-DL, Molhiv, Brain Tumor, and PPA.
% \begin{equation}
%     \text{Accuracy\textcolor{red}{pending!!!}} = \frac{1}{n} \sum_{i=1}^{n} \mathbb{1}\!\left[\hat{y}_i = y_i\right],
%     \label{eq:accuracy}
% \end{equation}
% where $\hat{y}_i$ and $y_i$ are the predicted and ground-truth class labels, respectively, and $\mathbb{1}[\cdot]$ is the indicator function.
% Higher accuracy indicates stronger discriminative capability.
% This metric is applied to the workloads of DMS, STEM-DL, Molhiv, and PPA.
%
For regression-based SciML workloads, we employ the coefficient of determination ($R^2$)~\cite{carpenter1960principles} to quantify how well the model explains the variance of the target variable: $R^2 = 1 - \frac{\sum_{i=1}^{n}(y_i - \hat{y}i)^2}{\sum{i=1}^{n}(y_i - \bar{y})^2}$.
% \begin{equation}
% R^2 = 1 - \frac{\sum_{i=1}^{n}(y_i - \hat{y}i)^2}{\sum{i=1}^{n}(y_i - \bar{y})^2},
% \label{r2}
% \end{equation}
where $\bar{y}$ denotes the mean of the ground-truth outputs.
An $R^2$ value closer to~1 indicates stronger predictive fidelity.
This metric is used for regression workloads, including CFD, PureMD, Fluid, and Synthetic.
For continuous-valued prediction tasks emphasizing absolute accuracy, we adopt the mean squared error (MSE):$\mathrm{MSE} = \frac{1}{n} \sum_{i=1}^{n} (y_i - \hat{y}_i)^2$
% \begin{equation}
% \mathrm{MSE} = \frac{1}{n} \sum_{i=1}^{n} (y_i - \hat{y}_i)^2,
% \label{mse}
% \end{equation}
where lower values indicate higher precision.
MSE is applied to workloads such as CosmoFlow, EM-Denoise, SLSTR, and Optical.

As shown in Figure~\ref{Accuracy overall}, \fram generally preserves the predictive accuracy of the baseline SciML models and even improves performance in several cases.
For regression workloads such as CosmoFlow, Optical, Synthetic, and PureMD, \fram achieves modest accuracy gains while substantially reducing model parameters.
Minor yet acceptable accuracy drops are observed in CFD (2.68\%) and Fluid (0.52\%).
For classification workloads such as Brain Tumor and STEM-DL, the decrease is limited to 0.69–1.59\%.
Notably, \fram alleviates overfitting in highly parameterized models; for instance, in DMS, it yields a clear accuracy improvement by regularizing redundant weights during training.

\ding{184} \textit{Effectiveness of Parameter Optimization.}  
Figure~\ref{fig:parameter_reduction} shows the parameter sizes before and after applying \fram across all workloads. 
\fram significantly reduces the number of active parameters from 37.4\% (PPA) to 99.9\% (CFD) with an average reduction ratio of 71.7\%.  
These reductions result from adaptive gating that removes non-informative connections and dynamically reallocates model capacity during training.  
Importantly, the processed models remain structurally valid and executable, as verified by the consistency checking described in Section~\ref{sec:scd}.  

\ding{185} \textit{Floating-Point Operations (FLOPs).}  
FLOPs quantify computational overhead and reflect the reduction in operational complexity.  
Figure~\ref{fig:flops} compares FLOPs before and after applying \fram across all evaluated workloads.  
The baseline FLOPs range from 0.24 million (PureMD) to over 100 billion (STEM-DL), covering a wide variety of SciML architectures.  
Applying \fram consistently lowers computational demand, achieving reductions from 6.3\% (CosmoFlow) to 99.9\% (CFD), with an average decrease of 69.6\%.  
These results confirm that \fram effectively minimizes redundant computations by jointly optimizing feature relevance and parameter utilization, thereby enhancing both efficiency and scalability.

\begin{table}[htbp]
\centering

\caption{Comparison with SOTA pruning methods}
\label{comparison}
\scriptsize
\renewcommand{\arraystretch}{1.3}
\begin{tabular}{l|cccc|c}
\hline 
\textbf{Metrics} & \textbf{OTO} & \textbf{L1} & \textbf{NEURAL} & \textbf{DepGraph} & \textbf{\fram} \\
\hline
One-Shot & Yes & No & No & Yes & Yes \\
Compression & Yes & Yes & No & Yes & Yes \\
Adaptiveness & No & No & No & No & Yes \\
\hline 
Puremd(\%\textcolor{red}{$\uparrow$}) & 72.73\% & 76.3\% & 80.10\% & 76.35\% & 76.10\% \\
Puremd(R²\textcolor{red}{$\uparrow$}) & 0.9430 & 0.9969 & 0.9318 & 0.9937 & 0.9988 \\
 
CFD(\%\textcolor{red}{$\uparrow$}) & 99.10\% & 99.12\% & 99.10\% & 99.12\% & 99.10\% \\
CFD(R²\textcolor{red}{$\uparrow$}) & 0.5933 & 0.9136 & 0.9785 & 0.9747 & 0.9732 \\
 
EMD.(\%\textcolor{red}{$\uparrow$}) & 30.10\% & 39.30\% & 39.50\% & 39.33\% & 39.09\% \\
EMD.(MSE\textcolor{green}{$\downarrow$}) & 9.13e-3 & 0.0401 & 0.021 & 0.041 & 9.51e-3 \\
 
DMS.(\%\textcolor{red}{$\uparrow$}) & 91.00\% & 91.02\% & 91.09\% & 91.22\% & 90.92\% \\
DMS.(ACC\textcolor{red}{$\uparrow$}) & 67.35 & 75.94 & 72.06 & 71.11\% & 87.33 \\
 
Bra.(\%\textcolor{red}{$\uparrow$}) & 60.34\% & 59.98\% & 61.09\% & - & 60.75\% \\
Bra.(ACC\textcolor{red}{$\uparrow$}) & 78.41 & 80.94 & 79.06 & - & 81.77 \\
\hline 
\end{tabular}
\begin{flushleft}
\footnotesize{
Note: `$\textcolor{red}{\uparrow}$' indicates that higher is better; `$\textcolor{green}{\downarrow}$' indicates that lower is better; `-' indicates that the latest version can't support this application.
}
\end{flushleft}
\vspace{-6mm}
\end{table}

\subsection{Comparison with Other SoTA Model Compression Methods}
We compare \fram against four representative frameworks from the four categories of state-of-the-art model optimization methods discussed in Section~\ref{sec:background}-B:
\textit{(1) Only Train Once (OTO)}~\cite{chen2021only},
\textit{(2) L1-norm regularization (L1)}~\cite{li2016pruning},
\textit{(3) NEURAL}~\cite{liu2017learning}, and
\textit{(4) DepGraph}~\cite{fang2023depgraph}.
We evaluate all methods across five representative workloads, including PureMD, CFD, EM-Denoise, DMS, and Brain Tumor. These workloads cover the five-layer types described in Section~\ref{incorporation}.
We assess performance using both the parameter reduction ratio, listed in the first row, and accuracy metrics (\texttt{$R^2$}, classification accuracy(\texttt{Acc}), and \texttt{MSE} depending on each task’s objective) shown in the second row of each evaluated workload.

% Table~\ref{comparison} presents the performance comparison. The results demonstrate that \fram consistently achieves superior or competitive performance across all benchmarks while being the only method offering adaptive compression capabilities.

% \fram\ exhibits several key strengths compared to existing methods. First, as shown in the DMS and Brain Tumor tasks, \fram\ achieves substantial accuracy improvements, reaching 87.33\% and 81.77\% respectively—outperforming all baselines by significant margins (e.g., 11.39 percentage points over L1 on DMS). Second, \fram\ demonstrates exceptional performance on regression tasks, achieving the highest $R^2$ score of 0.9988 on Puremd, indicating superior predictive quality. Third, \fram\ maintains competitive compression ratios while delivering better quality metrics, as evidenced by its low MSE of 9.51e-3 on EMDenoise, comparable to OTO's 9.13e-3 but with more balanced overall performance. 
As shown in Table~\ref{comparison}, \fram performs consistently well across all evaluated workloads and metrics.
Compared with OTO, L1, NEURAL, and DepGraph, \fram achieves competitive or higher accuracy while maintaining strong compression efficiency.
For regression tasks such as PureMD and CFD, \fram reaches the highest $R^2$ scores (0.9988 and 0.9732) with similar or fewer parameters than other methods.
In the EM-Denoise task, it records the lowest mean squared error ($9.51 \times 10^{-3}$), indicating more precise predictions.
For classification tasks such as DMS and Brain Tumor, \fram also achieves high accuracy (87.33\% and 81.77\%), comparable to or better than existing methods.
These results show that \fram can effectively balance model size and prediction quality, providing a simple and adaptive approach that works well across different SciML architectures.

\subsection{Feature Selection Analysis }
\begin{figure}[!ht]
    \centering
    % \vspace{-10pt}
    \includegraphics[width=0.45\textwidth]{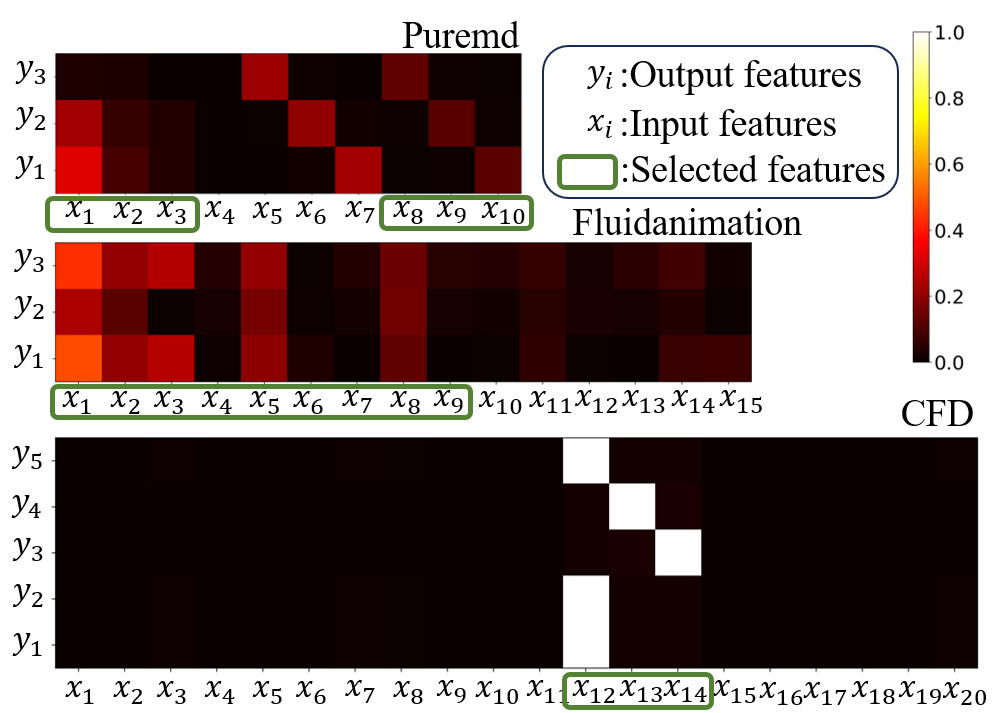}
    \caption{Showcase I: Feature selection efficiency in three regression workloads (CFD, PureMD, and Fluid).}
    
    \label{fssurrogate}
    \vspace{-1.0em}
\end{figure}

We evaluate \fram’s feature selection capability using three representative surrogate models, i.e., CFD, PureMD, and Fluid.
Each SciML model is trained on a distinct physical simulation (Figure~\ref{fssurrogate}), where the input–output data dependencies are analyzed and identified by a compiler-based framework~\cite{fink2024hpac}.
These tasks consist of numerical input arrays with complex inter-feature relationships, making them well-suited for assessing \fram’s effectiveness in selecting the most informative and relevant features.

To quantify feature relevance, we use the \emph{Pearson correlation coefficient} to measure the linear relationship between each input variable $\mathbf{x}$ and output variable $\mathbf{y}$:
\begin{equation}
\rho(\mathbf{x},\mathbf{y}) = \frac{\mathrm{Cov}(\mathbf{x}, \mathbf{y})}{\sigma(\mathbf{x}), \sigma(\mathbf{y})},
\label{pearson}
\end{equation}
where $\mathrm{Cov}$ denotes covariance and $\sigma(\cdot)$ represents the standard deviation.
A larger absolute value of $\rho$ indicates a stronger linear dependency between the input and output variables.

As shown in Figure~\ref{fssurrogate}, \fram consistently identifies the input dimensions that exhibit the highest correlations with their corresponding outputs.
For instance, in the CFD model, \fram automatically selects three key input features ${x_{12}, x_{13}, x_{14}}$ that are most correlated with the target outputs ${y_{1}, ..., y_{5}}$, while suppressing inputs with negligible correlation ($|\rho| \approx 0$).
Similarly, in the PureMD and Fluid models, the selected features align with the dominant physical variables governing system behavior, demonstrating the interpretability and reliability of \fram’s gating-based feature optimization.
These results confirm that \fram effectively identifies and retains scientifically meaningful features while eliminating redundant or weakly relevant inputs.

\subsection{Peak Memory Usage and Energy Saving}
\begin{figure}[htbp]
    \centering
    % \includegraphics[width=0.49\textwidth]{updated_figures/energy-v3.pdf}
    % \caption{Normalized energy consumption}
    % \label{fig:energy}

    \begin{subfigure}[b]{0.492\textwidth}
        \includegraphics[width=\textwidth]{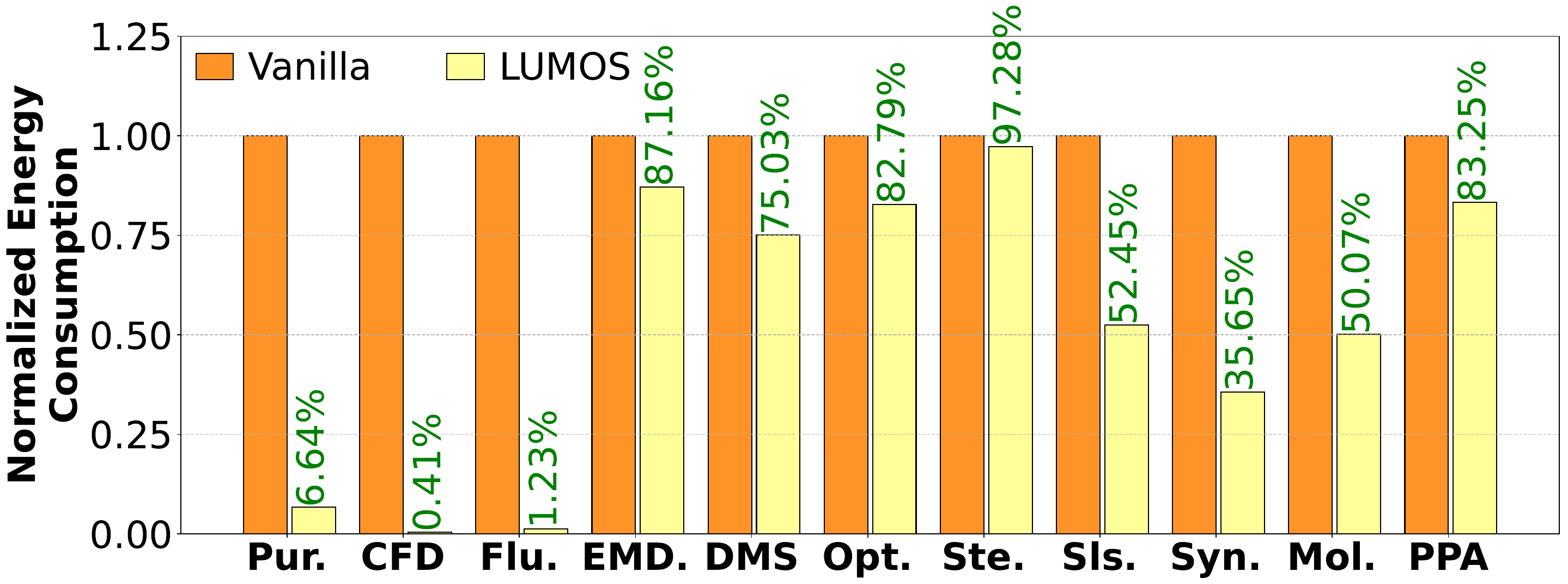}
        \caption{Energy consumption}
        \label{fig:energy}
    \end{subfigure}
    
    \vspace{0.5em} % 控制上下间距
    
    % 第二行
    \begin{subfigure}[b]{0.492\textwidth}
        \includegraphics[width=\textwidth]{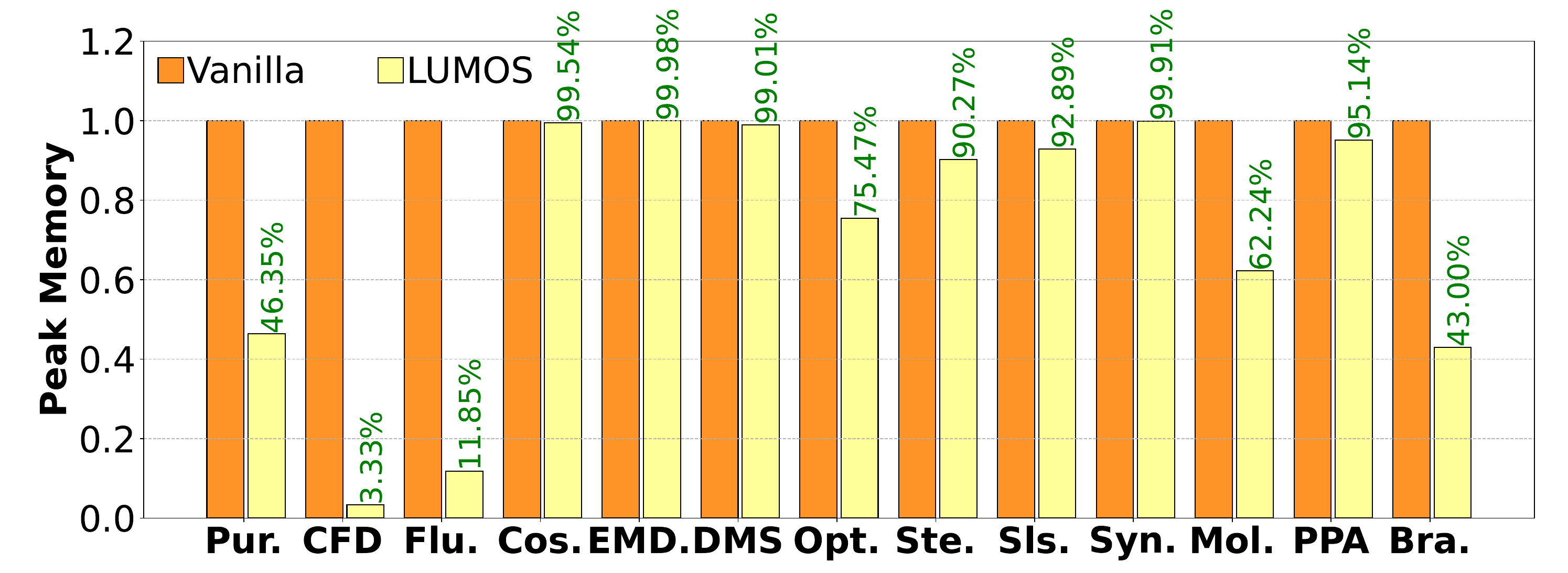}
        \caption{Memory usage}
        \label{fig:memory}
    \end{subfigure}
    \caption{Peak Memory Usage and Energy Saving.}
\vspace{-1.0em}
\end{figure}

\textit{Energy Consumption} quantifies power usage, providing an additional perspective on \fram’s operational and environmental efficiency.
Figure~\ref{fig:energy} presents the energy measurements obtained using a state-of-the-art power profiling tool~\cite{tu2023unveiling}.
On average, \fram reduces energy consumption by 50.7\% across all evaluated workloads.
For compact, FC layer–dominant models such as PureMD, CFD, and Fluid, energy savings exceed 90\%, highlighting the strong correlation between reduced computational cost and lower power demand.
Moderate yet consistent reductions of 30–60\% are also observed in image- and graph-based workloads, including SLSTR, Synthetic, and Molhiv.

\textit{Peak Memory Usage} measures the maximum memory consumption during execution.
Figure~\ref{fig:memory} shows the reduction in peak memory achieved by \fram.
Across all workloads, \fram reduces memory usage by up to 96.7\%.
The largest savings are observed in compact models such as PureMD, CFD, and Fluid, with reductions of 555~MB, 988~MB, and 914~MB, respectively.
In larger convolutional architectures, memory savings are smaller because intermediate activation tensors in CONV layers dominate memory consumption and are less affected by feature or parameter optimization.
Even so, \fram effectively reduces both activation and weight footprints, allowing larger batch sizes and higher throughput during training and inference.

\subsection{Scalability and Training Convergence}
\begin{figure}[!ht]
    \centering
    % \vspace{-10pt}
    \includegraphics[width=0.45\textwidth]{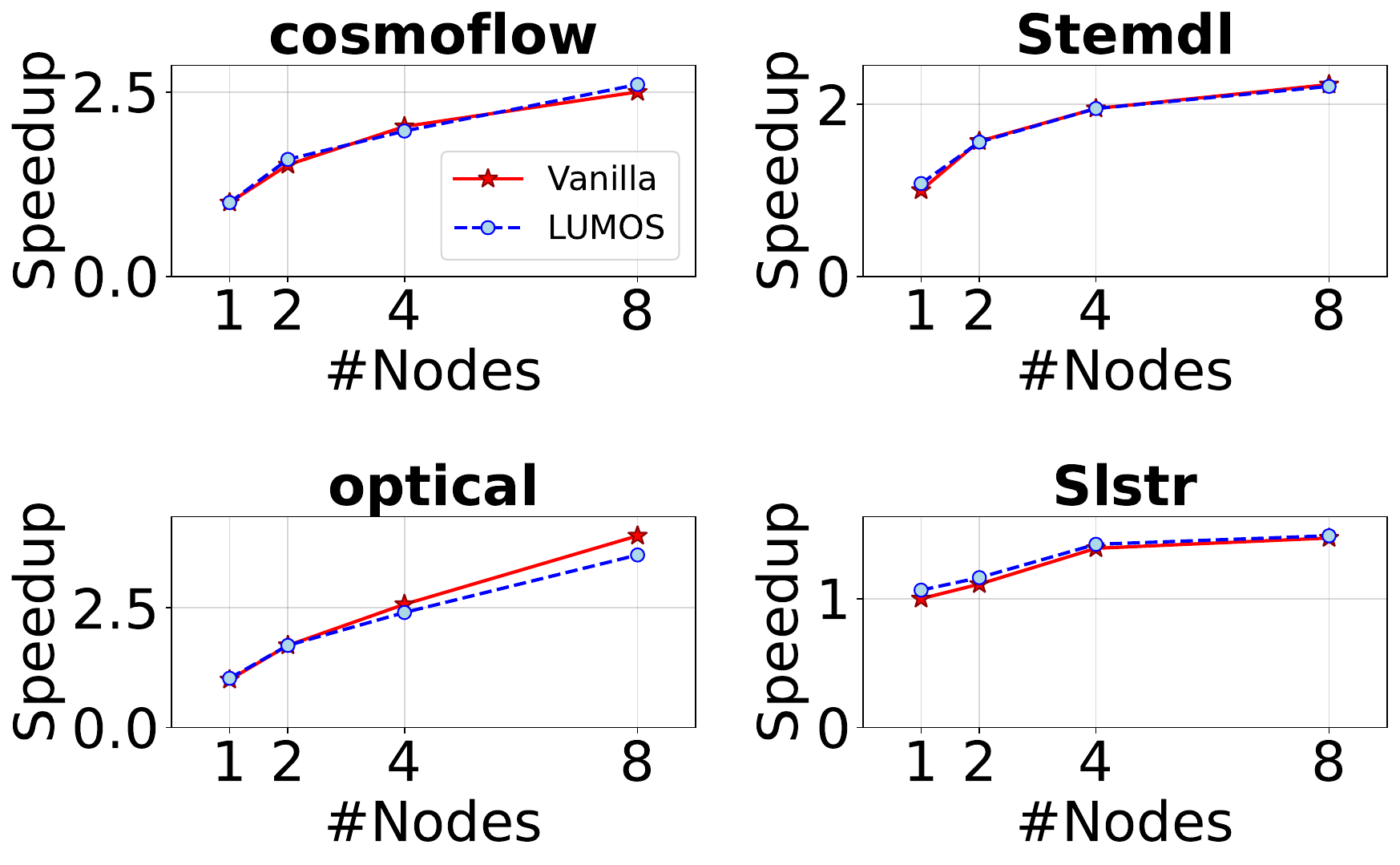}
    \caption{Scale \fram up to 8 GPUs using DDP training.}
    \label{scaling}
\end{figure}

\textit{Scalability Analysis.} Figure~\ref{scaling} illustrates the scalability of \fram compared to the original training approach across four different models: Cosmoflow, Stemdl, Slstr, and Optical.   All experiments were conducted using Distributed Data Parallel (DDP) training on up to 8 NVIDIA V100 GPUs (32GB each). Each model was trained for 200 epochs to ensure convergence, with batch size per GPU kept constant across different node counts for fair comparison.

The results show that \fram-integrated training has nearly the same scalability as vanilla training.
\fram introduces no observable scalability bottlenecks, as it adds lightweight gating parameters to each neuron without modifying the model’s architecture or communication pattern.
On average, these additional gates account for only about 3\% of the total parameters across the 13 evaluated models, imposing negligible overhead and remaining fully compatible with standard DDP training.

\begin{figure}[!ht]
    \centering
    % \vspace{-10pt}
    \includegraphics[width=0.45\textwidth]{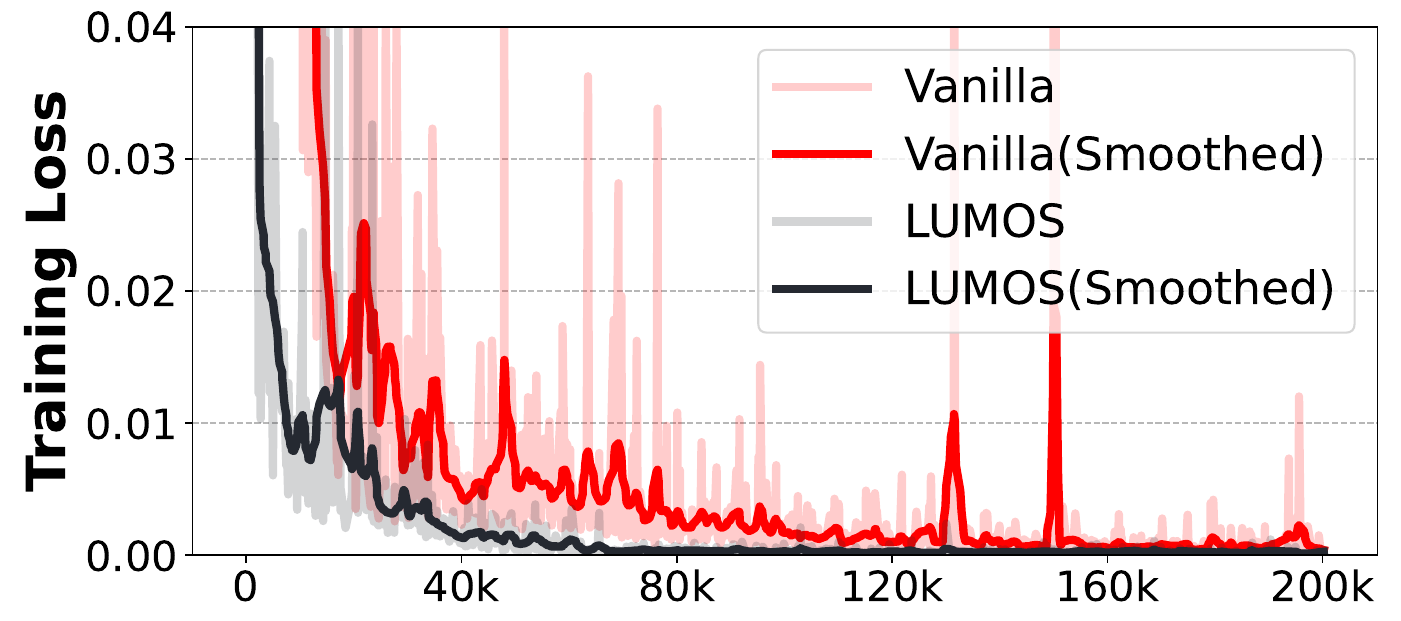}
    \caption{Showcase II: Training loss over time for the baseline and \fram-augmented models on the PureMD workload. \fram achieves faster convergence, lower final loss, and greater training stability compared with the baseline training process.}
    \label{loss-compare}
    % \vspace{-1mm}
\end{figure}

\textit{Training Convergence.}
Figure~\ref{loss-compare} shows the convergence behavior of the baseline and \fram-optimized models over 200{,}000 training steps.
Throughout training, \fram consistently achieves lower loss values than the baseline.
The smoothed optimized loss curve (black) descends more rapidly and stabilizes at a lower final value than the smoothed baseline curve (red), demonstrating faster convergence and improved training stability.

\begin{table}[!ht]
  \centering
  \caption{Training-time(minutes) overhead of integrating LUMOS.}
  \label{tab:lumos_overhead}
  \vspace{-0.3em}
  \small
  \begin{tabular}{lrrr}
    \toprule
    \textbf{Workload} & \textbf{Original} & \textbf{LUMOS} & \textbf{Overhead} \\
    \midrule
    cifar10  & 55  & 61  & +10.91\% \\
    DMSNet   & 80  & 82  & +2.50\%  \\
    PPA      & 128 & 137 & +7.03\%  \\
    Molhiv   & 43  & 41  & $-4.65\%$ \\
    Slstr    & 129 & 131 & +1.55\%  \\
    CFD      & 34  & 37  & +8.82\%  \\
    \bottomrule
  \end{tabular}
  \vspace{-0.8em}
\end{table}
\textit{Training Overhead.}
Table~\ref{tab:lumos_overhead} reports the training-time overhead introduced by LUMOS across representative workloads. Overall, the preliminary results indicate that LUMOS incurs only a modest overhead (below 11\%), which is expected: the semi-stochastic gates introduce a small fraction of additional learnable parameters and only lightweight, element-wise computations, while being optimized jointly within a single end-to-end training run. In contrast, typical pruning baselines (e.g., iterative magnitude pruning or multi-round structured pruning) often require repeated prune-and-retrain cycles, leading to substantially higher wall-clock costs. Taken together, LUMOS preserves near-vanilla training cost, consistent with the same tendency observed in our broader experiments.

\section{Conclusion}
\label{sec:conclusion}
\fram is a comprehensive end-to-end framework that unifies input feature selection, model sparsification, and structured model extraction for SciML workloads.
By leveraging semi-stochastic gating, \fram integrates feature selection and parameter optimization directly into the training process while remaining fully compatible with standard SciML layer types.
Designed to be hardware-agnostic and easy to integrate, \fram enables researchers to deploy efficient models in resource-constrained environments without compromising accuracy.
Extensive evaluations across 13 diverse scientific applications show that \fram significantly reduces FLOPs, inference latency, memory footprint, and energy consumption, providing a practical and scalable solution for next-generation scientific machine learning construction pipelines.

\section{Acknowledgment}

This work is supported by the U.S. National Science Foundation (2514351 and 2505118), and also supported by the U.S. Department of Energy, Office of Science, Advanced Scientific Computing Research (ASCR), under contracts \texttt{DE-AC02-06CH11357}. We thank the anonymous reviewers for their valuable feedback.

\bibliographystyle{unsrt} 
\bibliography{wdong, v-1}

\end{document}